\documentclass{ws-us}
\usepackage[sort,compress,super]{cite}
\usepackage{threeparttable}
\usepackage{graphicx}
\usepackage{multirow}
\usepackage{amsmath,amsfonts,amssymb}
\usepackage{tabularx}
\usepackage{url}
\usepackage[table,xcdraw]{xcolor}
\usepackage{algorithm}
\usepackage{adjustbox}
\usepackage{overpic}
\usepackage[noend]{algpseudocode}
\usepackage{changepage}
\usepackage{multirow}
\usepackage{colortbl}
\usepackage{threeparttable}
\usepackage{algorithm}
\usepackage{rotating}  
\usepackage[noend]{algpseudocode}
\usepackage{amsmath,amsfonts,amssymb}
\definecolor{Illusion}{rgb}{0.945,0.596,0.788}
\definecolor{MoonRaker}{rgb}{0.749,0.741,0.945}

\newlength{\extralength}
\newcommand{\overbar}[1]{\mkern 1.5mu\overline{\mkern-1.5mu#1\mkern-1.5mu}\mkern 1.5mu}
\begin{document}
\setlength{\pdfpagewidth}{8.5in}
\setlength{\pdfpageheight}{11in}
\catchline{0}{0}{2013}{}{}

\markboth{Otobong Jerome}{Kinematically Feasible Off-Road Global Planning}

\title{A Real-Time Framework for Intermediate Map Construction and Kinematically Feasible Off-Road Planning without OSM}

\author{Otobong Jerome$^\dagger$\footnote{Corresponding author}, Geesara Kulathunga$^{\ddagger,}$, Devitt Dmitry$^\dagger$, Eugene Murawjow$^\dagger$, Alexandr Klimchik$^\ddagger$}

\address{$^\ddagger$School of Computer Science, University of Lincoln, United Kingdom\\
E-mail: gkulathunga@lincoln.ac.uk, aklimchik@lincoln.ac.uk}

\address{$^\dagger$Center of Autonomous Technologies, Innopolis University, Innopolis, Russia\\
E-mail: o.jerome@innopolis.university, d.devitt@innopolis.university, eugene.murawjow@innopolis.university}

\maketitle

\begin{abstract}
Off-road environments present unique challenges for autonomous navigation due to their complex and unstructured nature. Traditional global path-planning methods, which typically aim to minimize path length and travel time, perform poorly on large-scale maps and fail to account for critical factors such as real-time performance, kinematic feasibility, and memory efficiency. This paper introduces a novel global path-planning method specifically designed for off-road environments, addressing these essential factors. The method begins by constructing an intermediate map within the pixel coordinate system, incorporating geographical features like off-road trails, waterways, restricted and passable areas, and trees. The planning problem is then divided into three sub-problems: graph-based path planning, kinematic feasibility checking, and path smoothing. This approach effectively meets real-time performance requirements while ensuring kinematic feasibility and efficient memory use. The method was tested in various off-road environments with large-scale maps up to several square kilometers in size, successfully identifying feasible paths in an average of 1.5 seconds and utilizing approximately 1.5GB of memory under extreme conditions. The proposed framework is versatile and applicable to a wide range of off-road autonomous navigation tasks, including search and rescue missions and agricultural operations.
\end{abstract}


\begin{multicols}{2}

\section{Introduction}
Off-road navigation is essential for a wide range of applications, including search and rescue missions, agricultural and forestry operations, and construction and surveying projects. Unlike on-road navigation, off-road navigation often involves traversing uncharted and challenging terrains. To manage these complexities, hierarchical path-planning methods that integrate both global and local path planning have become the preferred approach\cite{zhang2023global}.  

However, each level of path planning presents its own set of challenges. Local path planning in off-road environments must contend with obstacles such as terrain elevation changes (e.g., bumps, ruts, and rocks), sudden environmental changes (e.g., rainfall), and the need for real-time replanning\cite{lee2023learning}.

Global path planning\cite{zhang2023global}, on the other hand, faces difficulties such as handling large-scale topographic maps, accurately incorporating geographical features, planning for longer horizons, and avoiding local minima. While local path planning focuses on ensuring smooth vehicle maneuvering and preventing abrupt movements, global path planning is concerned with generating long-range paths that guide the local planner in achieving its objectives.

In recent years, various global path-planning approaches have been proposed for off-road navigation. One such approach is an improved A-star algorithm designed for long-distance off-road path planning\cite{hong2021improved}, but it does not account for vehicle kinematic constraints. 

Another approach\cite{yin2023reliable} incorporates the vehicle's maximum achievable speed and vertical acceleration as constraints within an RRT*-based path planning framework to enhance safety. However, RRT* faces scalability and memory-related challenges when applied to large-scale maps\cite{yin2023path}. To address these issues, J. Yin et al.\cite{yin2023efficient} introduced ER-RRT*, an enhanced version of RRT*, but its probabilistic nature hinders real-time performance.

In local planning, receding horizon methods like Model Predictive Control (MPC) and its variants optimize trajectories over a finite horizon, accounting for vehicle dynamics and environmental constraints \cite{rahab2025, chavoshi2025}. 

Sampling-based approaches MPC variants like Model Predictive Path Integral (MPPI) \cite{hannes2025} and Genetic Algorithmic Kinodynamic (GAKD) planning \cite{jerome2025genetic} offer scalable trajectory optimization under uncertainty using parallel sampling. However, these methods demand significant computational resources and may struggle to ensure feasibility over long horizons in complex terrains.

Gradient-based optimization techniques, such as Sequential Quadratic Programming (SQP) \cite{adeyemi2025, tang2025collaborative}, Covariant Hamiltonian Optimization for Motion Planning (CHOMP) \cite{zucker2013chomp}, and Iterative Linear Quadratic Regulator (ILQR) \cite{cheng2024}, are widely used for trajectory refinement. These methods effectively enforce dynamic constraints, producing smooth, kinematically feasible paths.

Recent advances in deep reinforcement learning (Deep RL) enable policies that generalize across diverse terrains and scenarios. Combined with recurrent neural networks (RNNs), these policies capture temporal dependencies and adapt to dynamic environments \cite{yang2025rough,zhang2024recent,wang2020mobile}. However, Deep RL's data inefficiency and lack of safety guarantees limit its practical deployment, especially in real-time systems navigating unstructured environments, where it often serves as a local planner.

In this work, we aim to overcome these limitations by developing a global planner that prioritizes three critical factors: real-time performance, kinematic feasibility, and memory efficiency (Table~\ref{tab:intro}).

When solving off-road navigation tasks, OpenStreetMap (OSM)~\cite{bennett2010openstreetmap} and Digital Elevation Models (DEMs)~\cite {liu2023multilevel} become the primary source for extracting necessary geographical features for understanding and constructing trails for off-road navigation-related applications\cite{castro2023does}.

However, in some cases (like traversing unexplored environments or after a natural disaster) this data is not available or not actual any more. So, the main focus of this work is to investigate geographical areas that lack OSM and DEM geographical features.

Therefore, a topographic map that is a structured representation of the geographical features of an area, including elevation, relief, roads, and bodies of water for off-road path planning, is required to construct to solve the global path planning task efficiently. Therefore, this work mainly concentrates on global path planning ensuring the listed three critical factors for the areas with insufficient Open Street Map (OSM) data or lack of Digital Elevation Model (DEM) data. Hence, the \textbf{main contributions of this work} are as follows:

\begin{itemlist}
\item \textbf{Intermediate map construction framework:} A framework was proposed for preprocessing geographical features and constructing an \textbf{intermediate map in the pixel coordinate system}, specifically for areas \textbf{without available OpenStreetMap (OSM) data}. The framework consists of three main steps: (i) feature extraction from raw sources, (ii) transformation of features into the pixel coordinate system, and (iii) construction of an intermediate map suitable for planning.

\item \textbf{Real-time off-road global planner:} An \textbf{off-road global planner} was developed and validated to ensure \textbf{real-time performance}, \textbf{kinematic feasibility}, and \textbf{low memory utilization}. This was achieved through four key strategies: (i) dynamic map slicing to limit computational load, (ii) on-the-fly distance map computation, (iii) selective application of kinematic constraints to critical path segments, and (iv) smoothing to improve path length and curvature.
\end{itemlist}

They enable the solution of global path-planning navigation tasks in off-road environments devoid of geographical features.

\begin{table*}[ht!]
\centering
\tbl{Comparative analysis of off-road global path planning techniques with the proposed technique.\label{tab:intro}}
{\begin{threeparttable}
\small
\begin{tabular}{|l|l|l|l|l|}
\hline
Approach & Terrain extraction & Vehicle modeling & \begin{tabular}[c]{@{}l@{}}User-\\ defined \\ areas\end{tabular} & Performance \\ \hline
\begin{tabular}[c]{@{}l@{}}Astar with slope \\ of the terrain \cite{hong2021improved}\end{tabular} & DEM and OSM & \begin{tabular}[c]{@{}l@{}}predefined slope \\ constraint\end{tabular} & no & not provided \\ \hline
\begin{tabular}[c]{@{}l@{}}Astar with heuristic\\ cost based on the\\ mobility cost map \cite{hua2022global}\end{tabular} & \begin{tabular}[c]{@{}l@{}}DEM, OSM, and\\ land use and soil \\ type distribution\\ data\end{tabular} & \begin{tabular}[c]{@{}l@{}}the proposed fuzzy\\ mobility cost \\ to incorporate vehicle\\ kinematics\end{tabular} & no & \begin{tabular}[c]{@{}l@{}}computationally\\ expensive\end{tabular} \\ \hline
\begin{tabular}[c]{@{}l@{}}A reliability-based \\ mission planning \\ method based on \\ improved RRT* \cite{yin2023reliable}\end{tabular} & soil and slope map & \begin{tabular}[c]{@{}l@{}}an adaptive surrogate \\ model and a dynamic\\ ensemble-based \\ dynamic surrogate \\ modelling\end{tabular} & no & \begin{tabular}[c]{@{}l@{}}computationally \\ expensive and \\ high memory \\ utilization\end{tabular} \\ \hline
ER-RRT* \cite{yin2023efficient} & soil and slope map & \begin{tabular}[c]{@{}l@{}}a surrogate model for \\ vehicle mobility \\ prediction using Gaussian \\ Process Regression (GPR)\end{tabular} & no & \begin{tabular}[c]{@{}l@{}}computationally \\ expensive \\ and high memory \\ utilization\end{tabular} \\ \hline
\begin{tabular}[c]{@{}l@{}}The proposed \\ technique\end{tabular} & \begin{tabular}[c]{@{}l@{}}layered map of the \\ environment: \\ bodies of water,\\ off-road trails,\\ rivers, and slopes\end{tabular} & \begin{tabular}[c]{@{}l@{}}consider the provided\\ kinematical model within\\ specified areas only\end{tabular} & yes & \begin{tabular}[c]{@{}l@{}}computationally \\ efficient \\ and memory\\ -efficient\end{tabular} \\ \hline
\end{tabular}
\begin{tablenotes}
  \item[1] KNODE: Knowledge-based Neural Ordinary Differential Equations to augment a model, DEM: Digital Elevation Model, OSM: OpenStreetMap
\end{tablenotes}
\end{threeparttable}}
\end{table*}

\section{Related Work}
In recent years, many scholars~\cite{rastgoftar2018data, hong2021improved, chu2012local, JEROME2025106172, jerome2025meshnav} have focused on global and local path planning for off-road navigation-related tasks. Literature reviews have been carried out for off-road global path planning, as this is the main focus of this work. The objective of a global path planner can be manifold~\cite{wang2018off, liu2017global}, e.g., finding the shortest path, the path with the least energy consumption, or the fastest path.

To achieve these objectives, a detailed topographic map must be constructed before planning. Most frameworks rely on OpenStreetMap (OSM), an open geographic database that contains a variety of geographical features, to build a topographic map~\cite{borkowska2023openstreetmap}. Contrary to most frameworks, off-road GPS applications such as BackCountry Navigator, Gaia GPS, and AllTrails use their own topographic maps. 

These applications also use user-contributed data to improve existing topographic maps. Digital Elevation Models (DEMs)~\cite {liu2023multilevel} is another source that can be used to construct topographic maps, in addition to OpenStreetMap. DEMs represent geographical features, e.g., mountains, valleys, and bodies of water, with elevation. DEMs can also be used to estimate the slope and aspect of the terrain. For example, triangular mesh planning~\cite{liu2022reliability} is a way of path planning on a surface that uses a DEM for constructing a topographic map.  

However, mesh planning is computationally intensive, which limits its use as a global off-road path planner in many cases. If the necessary geographical features cannot be extracted from OpenStreetMap or DEM, a topographic map must be constructed. In such cases, a GPS device with a camera or LiDAR can be attached to an aerial vehicle to measure the latitude and longitude of the geographical area, as well as collect images or point clouds~\cite{zhang2023global}. These data can then be used to construct a topographic map of the area. This approach was utilized in the proposed framework.

There are a number of different algorithms that can be used for global path planning, each with its own advantages and disadvantages. Off-road global path planning faces more computational and memory-related challenges than on-road global path planning. In theory, Dijkstra's algorithm~\cite{goulet2023high}, A* \cite{kormoczi2023comparison}, Jump Point Search (JPS)~\cite{liu2017planning}, Probabilistic roadmaps (PRMs)~\cite{hongqing2021probabilistic}, and Rapidly exploring random trees (RRTs)~\cite{hwan2011anytime} or other path planning algorithms can be utilized.

However, sampling-based approaches such as RRTs and PRMs make it very difficult to obtain real-time replanning capabilities~\cite{liu2023specification} due to computational and memory-related challenges. Deterministic heuristic-based path planning algorithms can be optimal choices, but only after carefully structuring the construction of the graph, scaling the graph search when the graph is large, and pre-processing the topographic map in a way that makes it easier to build and process the graph. 

In addition to A*, JPS, and RRT*, other algorithms have been explored for off-road global path planning. For instance, D* Lite* ~\cite{koenig2002d} is an incremental search algorithm that efficiently replans paths when the environment changes, making it suitable for dynamic off-road navigation. 

Fast Marching Trees (FMT*)~\cite{janson2015fast} is a sampling-based planner that efficiently explores the configuration space by connecting sampled points in a way that minimizes path cost. However, FMT* can struggle in high-dimensional spaces or complex terrain where a large number of samples are required. 

Reinforcement Learning (RL)-based approaches ~\cite{wang2020mobile} have also gained interest, leveraging data-driven policies to adapt to diverse and unstructured environments. While RL methods can generalize well across different terrains, they often require extensive training data and computational resources, making them less practical for real-time applications. Given these considerations, the proposed framework focuses on heuristic-based and deterministic approaches while acknowledging the potential benefits of these alternative methods for future extensions.  

The choice of a graph-based search algorithm depends on a number of factors, including the complexity of the terrain, the accuracy of the topographic map, and the available computational resources and time.

For example, if the topographic map consists of off-road trails, it may be necessary to prioritize off-road trail planning followed by off-road planning. Dijkstra's algorithm can be used for planning through the off-road trails because distance information is available, which provides the shortest path between two points in a graph. A* or JPS can then be used for off-road planning because these algorithms use defined heuristic functions to guide the search.

\begin{figure*}[ht!]
    \centering
    \includegraphics[width=1\linewidth]{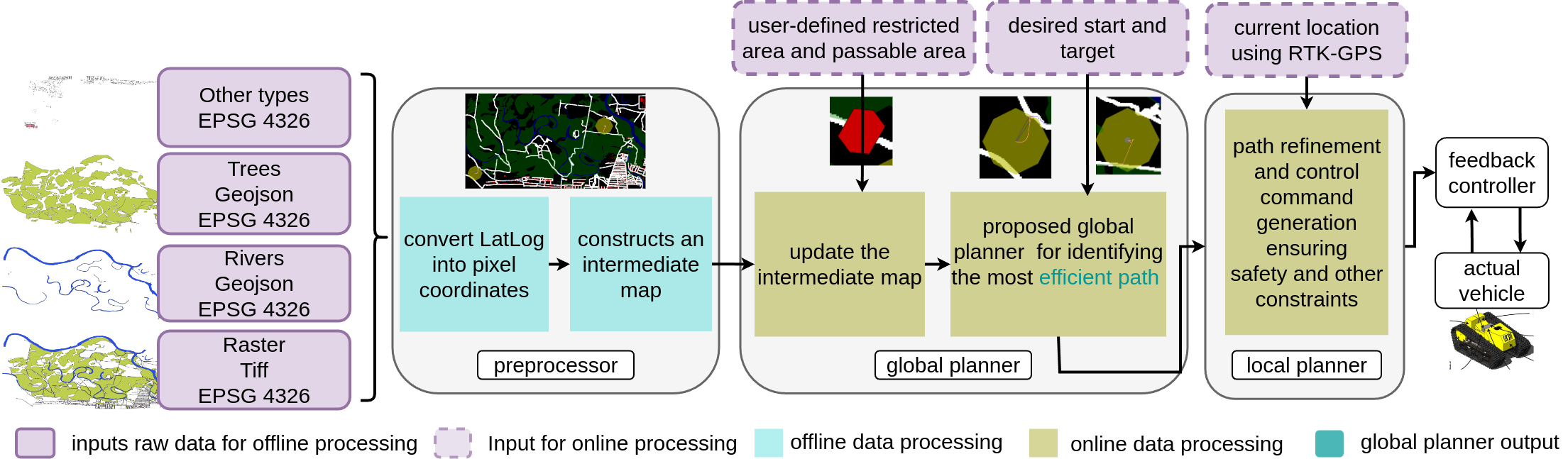}
    \caption{The proposed framework for off-road navigation. The initial step involves gathering essential raw data that encompasses the environment's features, including trees, buildings, bodies of water, and off-road trails. Once the raw data is preprocessed, an intermediate map is constructed as a preprocessing task. To accommodate user preferences and safety considerations, users are empowered to define restrictive and passable areas. The proposed global path planner leverages the constructed intermediate map and user-defined areas to generate real-time paths that adhere to kinematic constraints. Afterwards, we can integrate any off-road local planner for vehicle maneuvering.}
    \label{system_oveview}
\end{figure*}

\section{Methodology}
Most existing off-road navigation systems depend on OSM or DEMs. These datasets often lack coverage in remote or unexplored areas. The proposed framework (Fig.\ref{system_oveview}) tackles the challenge of navigating in such geographical terrains.  

To achieve this, we mounted an RTK-GPS, a camera, and a Livox Avia LiDAR \cite{livox} sensor on a quadcopter (a modified version of T-Drones-M690B)\cite{px4_lidar} (Fig.\ref{fig:data_collection}) and flew it over the target terrain. While traversing the area, we recorded the current pose in the EPSG 4326 coordinate system and captured synchronized images and point clouds along the pass.

We utilized LiDAR odometry and mapping (LOAM)\cite{lin2020loam} algorithm to construct an initial input map in EPSG 4326, providing a foundation for further analysis and refinement using QGIS tools. We employed Global Mapper\cite{globalmapper} software to extract intricate geographical features from the point cloud data. This included identifying off-road trails, bodies of water, and trees, providing a comprehensive understanding of the terrain's topography. The resulting topographic map is depicted in Fig.~\ref{fig:input_image}.  

For context, our focus in this study is primarily on planning, where the map serves as an input. We believe various mapping techniques can be employed depending on the application.
The accuracy and resolution of the generated map are significantly influenced by the specifications of the Livox Avia LiDAR sensor. Key factors such as point density, scanning range, and the sensor’s ability to penetrate vegetation directly affect the recognition of terrain features, obstacles, and edges.   

\vspace{1cm}
\begin{figurehere}
    \centering
    \begin{overpic}[width=1\linewidth]{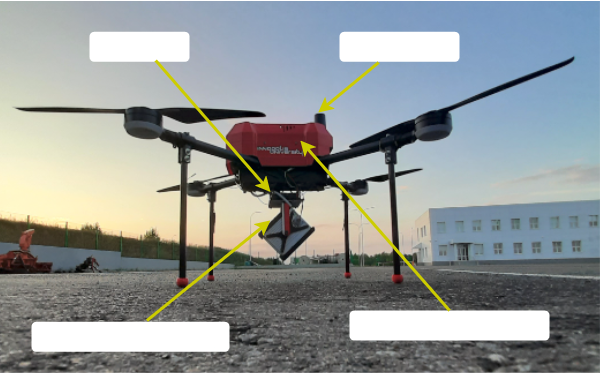}
    \put(15,53){Camera}
    \put(56.5,53){RTK-GPS}
    \put(5.5,5){Livox Avia LiDAR}
    \put(58.5,6.5){Nvidia Xavier Agx}
    \end{overpic}
     \caption{To map an uncharted terrain, we employed a T-Drone equipped with a Livox Avia LiDAR sensor.}
    \label{fig:data_collection}
\end{figurehere}

Higher point density improves the precision of object detection and classification, while the LiDAR’s field of view and angular resolution determine the level of detail captured in the environment.

Additionally, since LiDAR signals can be obstructed by dense vegetation, the ability to detect ground features in forested or overgrown regions may be limited. To mitigate these challenges, post-processing techniques such as ground filtering and vegetation classification were applied to refine the extracted topographic features.

These considerations are crucial for ensuring reliable terrain classification and accurate path planning in off-road environments. The proposed global path planning approach for off-road navigation is explained in the following sub-sections.

\subsection{Input data} \label{sec:input_data}  

The proposed off-road navigation system accepts several types of input files. First, georeferencing information is accepted as a GeoTIFF, which is a standardized open-source geographic metadata format allowing the integration of georeferencing information into TIFF (Tag Image File Format) images (Fig.\ref{fig:input_image}(a)).

The geographic metadata includes a transformation between the actual location in space and the pixel location in the image. Also, it includes horizontal and vertical datums, the Coordinate Reference System (CRS), ellipsoids and geoids, and map projection information.  

Second, the proposed off-road navigation system accepts geographical features (Fig.\ref{fig:input_image}(a,b)), such as roads, buildings, rivers, bodies of water, and user-restricted (or keep-out areas), as GeoJSON files. GeoJSON is an open standard geographical features interchange format that represents geographic features using simplified geometric shapes such as polygons, lines, points, and multi-polygons to a CRS. In the proposed framework, the World Geodetic System 1984 (WGS 84) [WGS84] reference frame is accepted as the desired CRS. 

\begin{figurehere}
    \centering
    \begin{overpic}[width=1\linewidth]{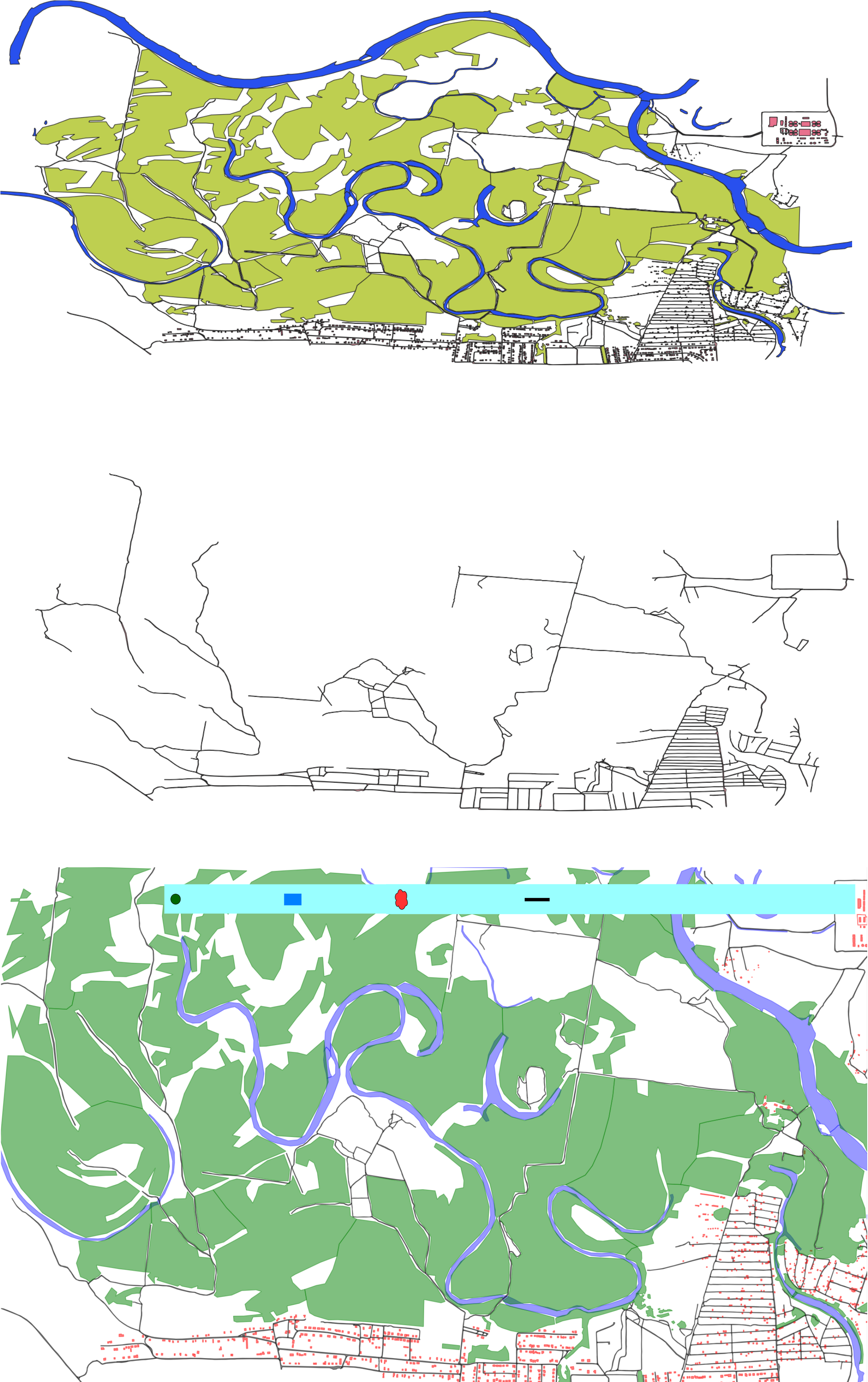}
    \put(0, 69.5){(a) Input map in the EPSG:4326 (WGS 84 -- World}
    \put(4, 67){Geodetic System 1984) coordinate system}
    \put(0, 39.5){(b) Off-road trails in the EPSG:4326 (WGS 84)}
    \put(4,37.5){coordinate system.}
    \put(0,-3){(c) Input map with geographical features in the pixel}
    \put(4,-6){coordinate system.}

    \put(14,34){Trees}
    \put(22,34){River}
    \put(29,34){Building}
    \put(40,34){Off-road trails}
    \end{overpic}
    \vspace{1cm}
    \caption{Georeferencing information and geographical features that are accepted as the input to the proposed off-road navigation framework. Initially, geographical features are converted into the pixel coordinate system and perform path planning and project planned path back to the original coordinate system (EPSG:4326)}
    \label{fig:input_image}
\end{figurehere}

The high-level concept behind the proposed global planner is to enable real-time, kinematically feasible, and memory-efficient off-road path planning by combining dynamic map slicing, on-the-fly distance map computation, and selective smoothing (Fig.~\ref{fig:flowchart}). The following sections provide a detailed explanation of each component.

\subsection{Intermediate map construction}

\begin{figurehere}
    \centering
    \begin{overpic}[width=1\linewidth]{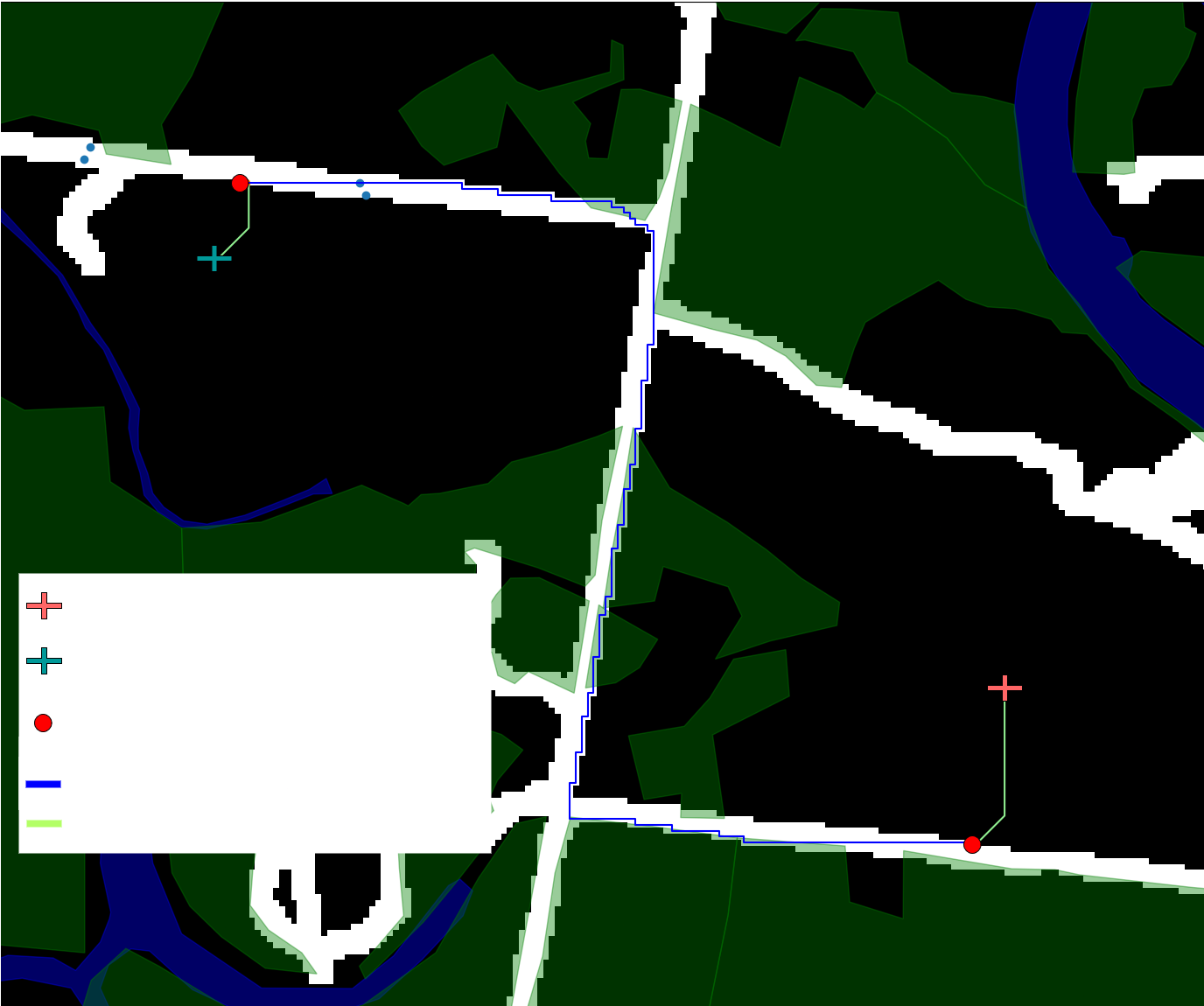}
    \put(6,32.5){Start pose}
    \put(6,27){Target pose}
    \put(6,22){Desired closest pose}
    \put(6,18){Planned path}
    \put(6,14){Off-road trails}
    \end{overpic}
    \caption{The proposed hybrid path planning algorithm leverages existing off-road trails and map data to determine a route between two positions. In this particular scenario, the start and target positions are located outside the off-road trails. However, the initially planned path is not the shortest possible. Algorithm~\ref{alg:find_closer_points} is designed to identify the shortest path. As shown in Fig.~\ref{fig:point_selections_clusters}, it successfully provides the shortest path for this example.} \label{fig:planned_path}
\end{figurehere} 

In the proposed system, we construct the intermediate map in the pixel coordinate system for two main reasons. First, it allows for efficient data storage and access, which is critical for fast path planning. 

Second, the pixel coordinate system offers higher spatial precision for local planning tasks than geographic coordinate systems like WGS84 (EPSG:4326). This is due to the fact that even small changes in latitude or longitude can translate into significant spatial shifts when projected into local coordinate frames such as UTM, unless handled carefully with scaling.  

In contrast, the pixel coordinate system provides consistent, resolution-aware spatial references, which are essential for precise collision checking and local trajectory refinement during path smoothing.

The conversion between geographic coordinates (latitude, longitude) and pixel coordinates is performed using the following transformation:

\begin{equation}
x\_{\text{pix}} = \frac{ \text{lon} - x\_{\text{origin}} }{\text{pixel\_width}}, \quad
y\_{\text{pix}} = \frac{ \text{lat} - y\_{\text{origin}} }{\text{pixel\_height}},
\end{equation} where $\text{lon}$ and $\text{lat}$ represent the longitude and latitude, and $x_{\text{origin}}$, $y_{\text{origin}}$ define the origin point of the image in EPSG:4326 coordinates. These origin values, along with $\text{pixel\_width}$ and $\text{pixel\_height}$, are extracted from the GeoTIFF's metadata using the GDAL library~\cite{warmerdam2008geospatial}.

After transforming all geographic features to pixel coordinates (Fig.~\ref{fig:input_image}(c)), each pixel is classified as traversable or non-traversable. This information is stored in the intermediate map. To reduce memory usage and accelerate access during path planning, the intermediate map is stored as a flattened 1D array of size \text{pixel\_width} $\times$ \text{pixel\_height}, instead of as a 2D grid.

The choice of pixel coordinate system plays a critical role not only in map preprocessing but also in downstream components of the path planning pipeline (Sections 3.3–3.7). For example, the high-resolution pixel-based representation enables precise obstacle avoidance during Hybrid A* planning and improves the fidelity of curvature-aware smoothing. This ensures that the generated paths are both accurate and feasible when mapped back to real-world coordinates.  

\vspace{1.5cm}
\begin{figurehere}
    \centering
    \begin{overpic}[width=1\linewidth]{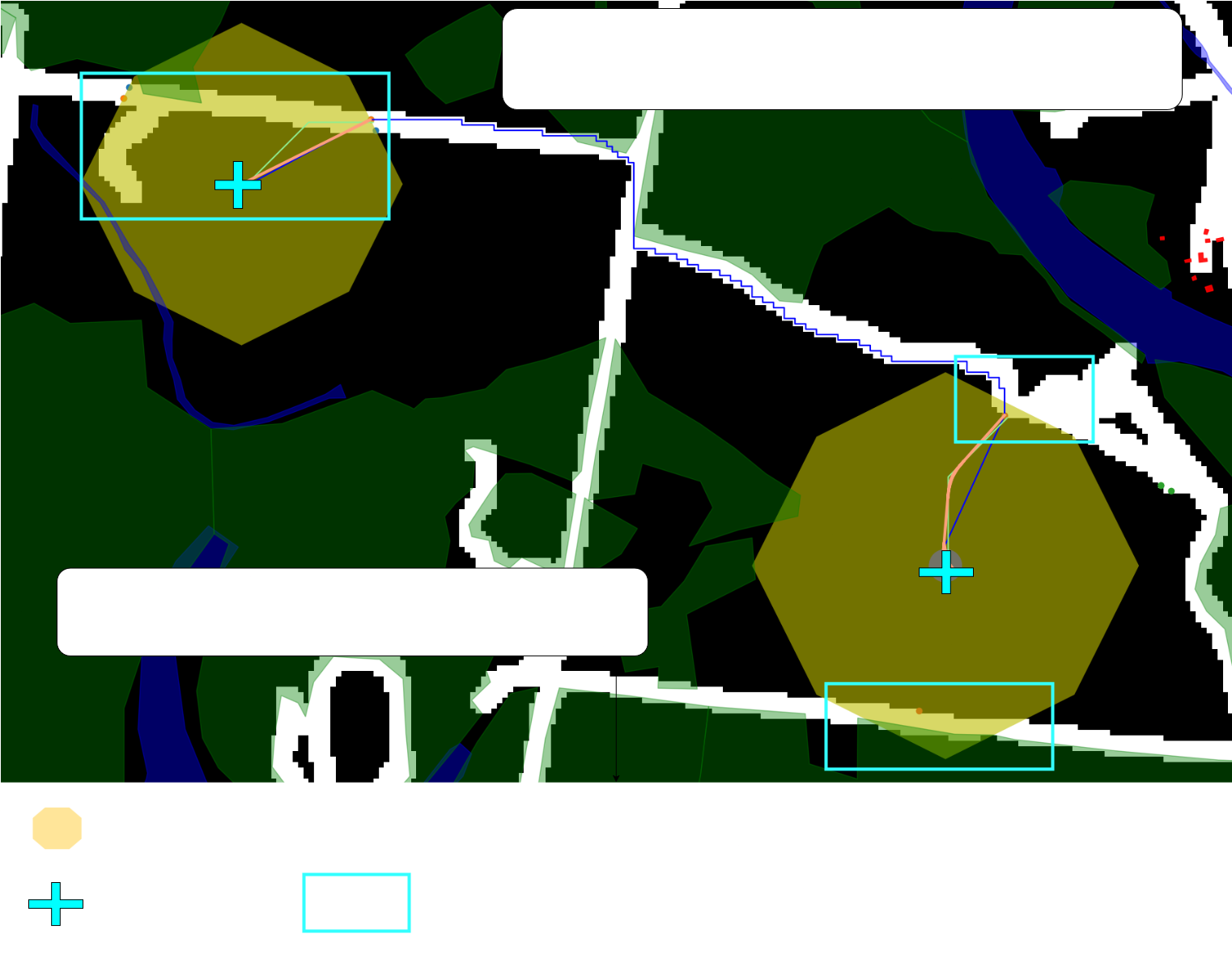}
        \put(40,74){ \small The goal-closest poses form a}
        \put(40,70){ \small single cluster.}

        \put(4,28.5){ \small Goal-proximate poses cluster}
        \put(4,25){ \small in multiple locations.}

         \put(7,9){\small Polygon for finding poses nearest to the goal 
         location.}
         \put(6,3){ \small Goal pose}
          \put(32,3){ \small Polygons and off-road trails overlap}
    \end{overpic}
        \vspace{0.5cm}

    \caption{Visual representation of Algorithm~\ref{alg:find_closer_points} for finding closer poses on the off-road trails to the goal pose}\label{fig:point_selections_clusters}
\end{figurehere}

\subsection{Find possible closer poses on the off-road trails with respect to start and target poses}\label{sec:road_map_path}
Given the start pose ($\mathbf{s_d}$) and the target pose ($\mathbf{t_d}$), it is quite challenging to obtain an optimal pose closer to the off-road trails. Consider the example illustrated in Fig.\ref{fig:planned_path}.

In this example, the closest pose to the off-road trails does not provide the shortest path between the $\mathbf{s_d}$  pose and the $\mathbf{t_d}$ pose. An approach (Algorithm \ref{alg:find_closer_points}) is proposed to correctly estimate the optimal closest pose to the goal location (start or target). 

\begin{figurehere}
    \centering
    \includegraphics[width=0.65\linewidth]{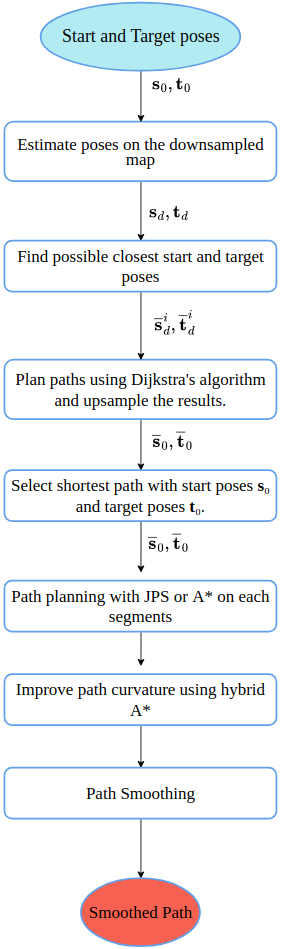}
    \caption{The high-level idea of the proposed global planner.}
    \label{fig:flowchart}
\end{figurehere}

\begin{algorithm}[H]
\caption{Finding Closest Valid Poses on Off-Road Trails to a Given Goal Pose}
\label{alg:find_closer_points}
\begin{algorithmic}[1]
\Statex \textbf{Inputs:}
\Statex \hspace{1em} $g_o$: \texttt{Pose} --- target goal pose
\Statex \hspace{1em} $poly_{md}, poly_{sd}$: \texttt{float} --- initial polygon size along main/side direction
\Statex \hspace{1em} $\overbar{poly}_{md}, \overbar{poly}_{sd}$: \texttt{float} --- maximum polygon size along main/side direction
\Statex \hspace{1em} $md_i, sd_i$: \texttt{float} --- polygon size increment for main/side direction
\Statex
\Statex \textbf{Outputs:}
\Statex \hspace{1em} $\bar{g}_o^i$, $i = 1, \dots, N$: \texttt{List[Pose]} --- list of selected nearby valid goal poses
\Statex \hrulefill

\While{$poly_{md} \leq \overbar{poly}_{md}$ \textbf{and} $poly_{sd} \leq \overbar{poly}_{sd}$}
    \State $poly \gets \text{getPolygon}(g_o, poly_{md}, poly_{sd})$
    \State $\text{covered\_points} \gets \text{rtreeCoveredBy}(poly)$
    \If{$\text{covered\_points} \ne \emptyset$}
        \State \textbf{goto} Step 12
    \EndIf
    \State $poly_{md} \gets poly_{md} + md_i$
    \State $poly_{sd} \gets poly_{sd} + sd_i$
\EndWhile

\Comment{Fallback: try intersecting points if no covered points were found}
\Statex

\While{$poly_{md} \leq \overbar{poly}_{md}$ \textbf{and} $poly_{sd} \leq \overbar{poly}_{sd}$}
    \State $poly \gets \text{getPolygon}(g_o, poly_{md}, poly_{sd})$
    \State $\text{intersected\_points} \gets \text{rtreeIntersectBy}(poly)$
    \If{$\text{intersected\_points} \ne \emptyset$}
        \If{$|\text{intersected\_points}| > 1$}
            \State \Return $\text{intersected\_points}$
        \EndIf
        \State \textbf{goto} Step 30
    \EndIf
    \State $poly_{md} \gets poly_{md} + md_i$
    \State $poly_{sd} \gets poly_{sd} + sd_i$
\EndWhile

\Statex
\Comment{If covered points exist, cluster them and select closest per cluster}

\If{$\text{covered\_points} \ne \emptyset$}
    \State $clusters \gets \text{DBSCAN}(\text{covered\_points})$
    \If{$|clusters| \geq 1$}
        \For{each $cluster_i$ in $clusters$}
            \State $\bar{g}_o^i \gets \text{getClosestPointToGoal}(g_o, cluster_i)$
        \EndFor
        \State \Return $\{ \bar{g}_o^i \mid i = 1,\dots,N \}$
    \EndIf
\EndIf

\State \Return $\{ g_o \}$ \Comment{Fallback: return original goal if no alternatives found}
\end{algorithmic}
\end{algorithm} 

According to Algorithm~\ref{alg:find_closer_points}, when a goal location is provided, all points within the relevant polygon are selected. These points are then grouped into clusters using the DBSCAN algorithm~\cite{hahsler2019dbscan} (as illustrated in Fig.\ref{fig:point_selections_clusters}), and one point from each cluster closest to the goal location is identified. 

If multiple points fall within the same cluster, the intersection between the polygon and the off-road trails is calculated, and the intersected points are selected as the closest points (Fig.\ref{fig:point_selections_clusters}).  

Finally, the Dijkstra algorithm is used to plan the path between the selected points and compute the distance between them, as detailed in Section~\ref{sec:distance_map}.

\subsection{Estimate distance map for Dijkstra algorithm}\label{sec:distance_map}
To reduce computational overhead and memory usage, the Dijkstra algorithm employs a downsampled map to generate the distance map. This downsampling factor, denoted as $d_f \in \mathbb{R}$, determines the level of granularity of the map, influencing the algorithm's efficiency and performance. For instance, if an original image with a resolution of 25,000 x 6,000 pixels is downsampled with $d_f=8$, the resulting image will have a resolution of 3,125 x 750 pixels. This downsampling process is applied to both the original intermediate map and the start and target poses, denoted as $\mathbf{s_o} \in \mathbb{R}^3$ and $\mathbf{t_o} \in \mathbb{R}^3$, respectively. The corresponding downsampled poses, $\mathbf{s_d} = \mathbf{s_o}/d_f \in \mathbb{R}^3$ and $\mathbf{t_d} = \mathbf{t_o}/d_f \in \mathbb{R}^3$, respectively, represent the positions in the 2D space, along with their heading angles $\theta$. Knowing the start pose $\mathbf{s_d}$, the distance map is calculated using wavefront propagation algorithm~\cite{xu2015fast}.

\begin{figure*}
    \centering
     \begin{overpic}[width=1\linewidth]{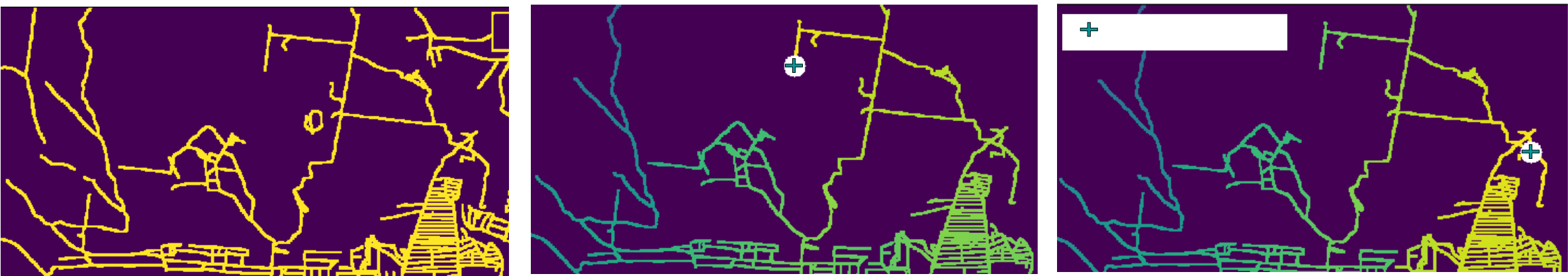}
        \put(-1,-2){ (a) Off-road trails }
        \put(33,-2){ (b)The distance map of off-road trails}
        \put(36,-4){from the given initial pose.}
        \put(67,-2){(c)Another distance map of off-road }
        \put(70,-4){trails from the given initial pose. }

        \put(71,15){Initial pose}
    \end{overpic}
    \vspace{0.5cm}
    \caption{Estimating the distance map for the Dijkstra algorithm using wavefront propagation on off-road trails.  The distance of traveling to any point on the off-road trails is calculated starting from the initial pose, which has a zero distance. The distance map is colored from yellow to purple, where yellow indicates the closest points to the initial pose and blue indicates the farthest points}
    \label{fig:distance_map}
\end{figure*} 

As the wavefront propagates away from the initial pose, the distance to the initial pose gets larger (Fig.\ref{fig:distance_map}). Once the distance map is constructed, it is up-sampled back to the original intermediate map size. A KD-tree~\cite{zhou2008real} is maintained to find the correspondence between the up-sampled and down-sampled off-road trails. Afterwards, the Dijkstra algorithm is used to plan a path within the off-road trails and calculate the distance between the start and target poses. The distance of a path is calculated by summing up the Euclidean distances between consecutive points of the planned path. The two optimal points, the start pose $\mathbf{\bar{s}_d}$ and the target pose $\mathbf{\bar{t}_d}$, are chosen based on the path with the minimum distance. Afterwards, the start pose ($\mathbf{\bar{s}_o} = \mathbf{\bar{s}_d}\cdot d_f$) and the target pose ($\mathbf{\bar{t}_o} = \mathbf{\bar{t}_d}\cdot d_f$) are up-sampled along with the path segment ($\mathbf{\bar{s}_o} \rightarrow \mathbf{\bar{t}_o}$). 

\subsection{Real-time path planning with kinematic constraints using JPS or A*}\label{sec:path_kino}  

After obtaining the road map-based path segment ($\mathbf{\bar{s}_o} \rightarrow \mathbf{\bar{t}_o}$) as described in Section~\ref{sec:distance_map}, two additional path segments, $\mathbf{s_o} \rightarrow \mathbf{\bar{s}_o}$ and $\mathbf{\bar{t}_o} \rightarrow \mathbf{t_o}$, are needed to complete the path from $\mathbf{s_o}$ to $\mathbf{t_o}$. To find these segments, the improved Jump Point Search (JPS)~\cite{liu2017planning} algorithm or the A* path planning algorithm is employed. Both the JPS and A* algorithms have been enhanced by incorporating vehicle parameters during the search for free space and smoothing the resulting path, as discussed in Section~\ref{sec:smoothing}. 

Initially, the A* or JPS algorithm is applied to find the intermediate points whose constructed path is not kinematically feasible since vehicle kinematics was not included when planning a path between $\mathbf{s_o}$ and $\mathbf{t_o}$. 

It is not possible to use kinodynamic path planning approaches\cite{jian2023long, greenberg2023gpu} directly due to the high computational cost. Typically, map resolution ranges from 10,000 to 30,000 pixels, corresponding to 1 to 6 kilometers in latitude-longitude form. Consequently, once intermediate points are identified, it is crucial to determine which segments of the path require re-planning to ensure the entire path's kinematic feasibility. To identify such path segments, we employed a bicycle model (Fig.\ref{fig:model_types}(a))\cite{polack2017kinematic}. 

\begin{figurehere}
    \centering
     \begin{overpic}[width=1\linewidth]{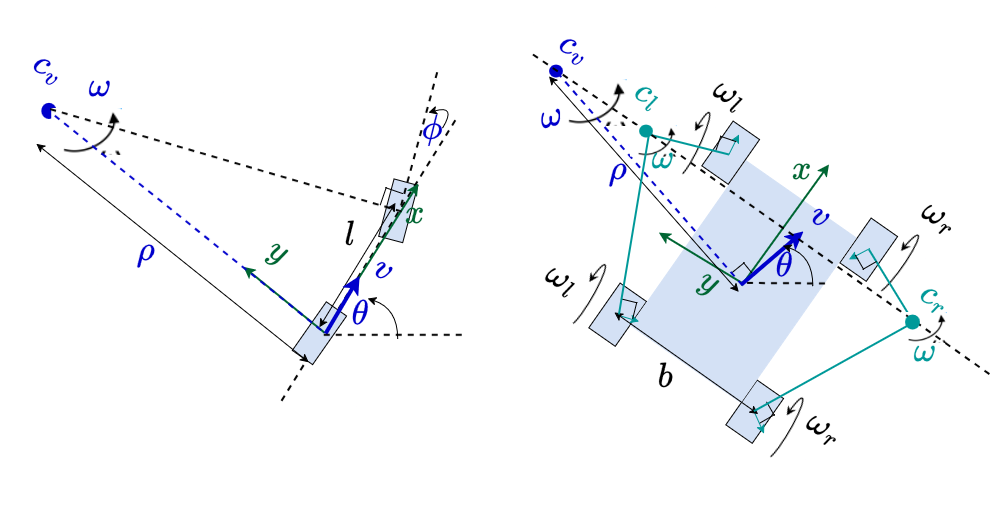}
        \put(5,1){  (a) Bicycle model }
        \put(53,1){ (b) Skid-steer model}
    \end{overpic}
    \vspace{0.5cm}
    \caption{Locating the instantaneous center of rotation ($c_v$) for a moving rigid body in a specific kinematic model.}
    \label{fig:model_types}
\end{figurehere}

However, the proposed approach applies to many ground vehicles' kinematic models, including Ackermann steering\cite{zhang2023energy}, differential drive\cite{wang2023differential}, and skid-steering\cite{baril2020evaluation} (Fig.~\ref{fig:model_types}(b)). Considering the bicycle model, the first step involves calculating the minimum turning radius. Let the speed $v$ and steering angle $\phi$ be control inputs $u_s$ and $u_{\phi}$, respectively, the state transition equation for the bicycle model is given by:  

\begin{equation}
    \begin{aligned}
        \dot{x} = u_s\cdot cos(\theta), \; 
        \dot{y} = u_s \cdot sin(\theta), \;
        \dot{\theta} = \frac{u_s}{l}\cdot tan(u_{\phi}),
    \end{aligned}
\end{equation} where minimum turning radius $\rho_{min} =l/tan{(\phi_{max})}$, $|u_{\phi}| < \phi_{max} < \pi/2$.   

\begin{figurehere}
    \centering
    \includegraphics[width=0.5\linewidth]{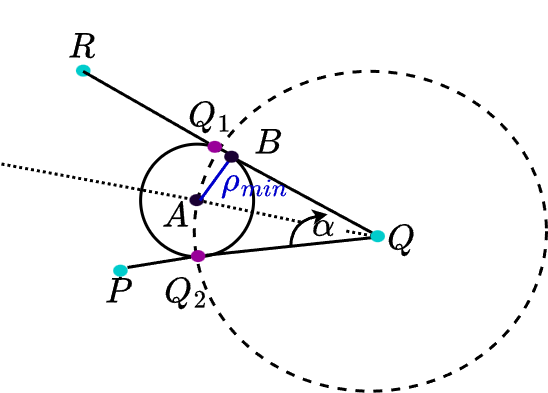}
    \caption{For the given path and model (Fig.\ref{fig:model_types}) with minimum turning radius ($\rho_{min}$) pinpoint segments requiring modification to satisfy kinematic feasibility constraints.}
    \label{fig:calculate_angle}
\end{figurehere} 

Afterwards, the angle $\alpha = \widehat{RQP}$ is measured. Then, the bisector of the angle is constructed (Fig.\ref{fig:calculate_angle}). Let $\rho_{min}$ be the radius, then the distance from the vertex to the circle center $QA$ is:
\begin{equation}
     s = \frac{\rho_{min}}{\sin \left( \frac{\alpha}{2} \right) + \epsilon}
\end{equation}
where $\epsilon$ is a small positive constant, e.g., $0.0001$ that prevents instability when $\alpha$ is close to zero. 

A vector of length $s$ is then numerically created along $QR$ and rotated by $\frac{\alpha}{2}$ to find point $A$. Hence, intermediate points $Q_1$ and $Q_2$ around Q can be determined. Afterwards, Hybrid A* is utilized for replanning the path between $Q_1$ and $Q_2$ only using a slice of the intermediate map.

\begin{figure*}
    \begin{overpic}[width=1\linewidth]{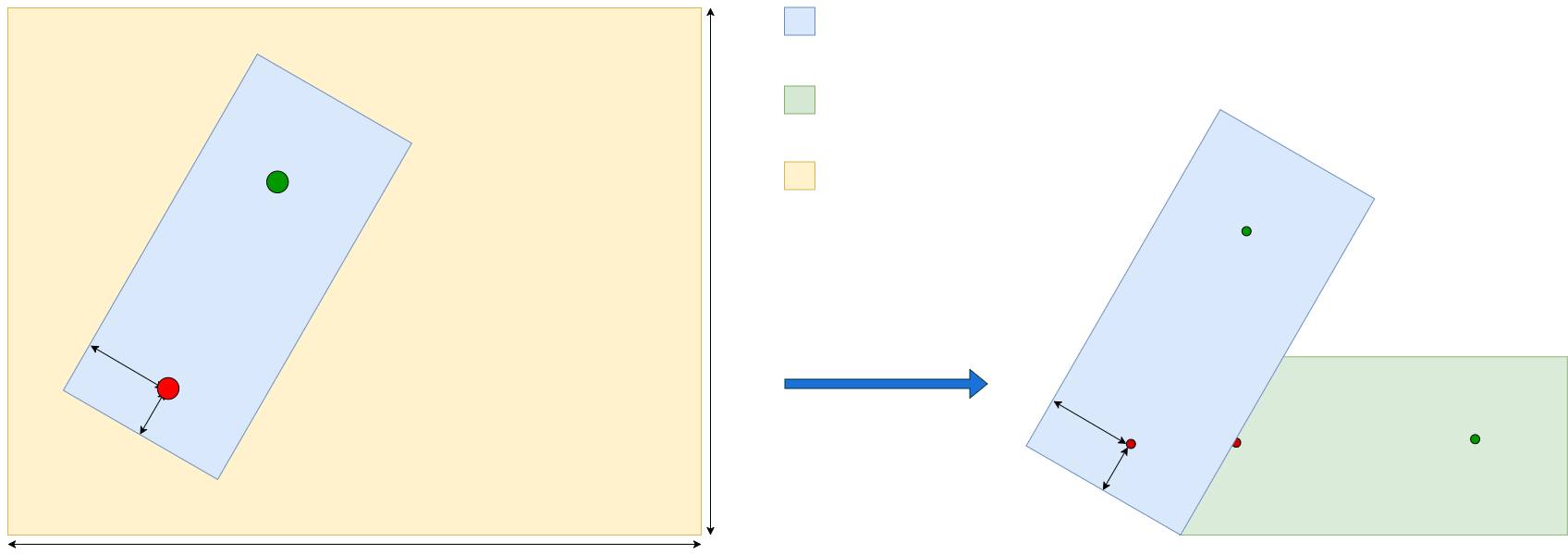}
        \put(52.5,33.5){Extracted intermediate map}
        \put(52.5,28.5){Extracted intermediate map (transformed)}
        \put(52.5,23.5){Original map}

        \put(50,18.5){$\mathbf{s}^i_o$}
        \put(52.5,18.5){Start pose}

        \put(50,13.5){$\mathbf{t}^i_o$}
        \put(52.5,13.5){Target pose}

        \put(20,-2){$m_x$}
        \put(46,19){$m_y$}

        \put(11.5,10.5){$\mathbf{s}^i_o$}
        \put(18.5,23.5){$\mathbf{t}^i_o$}

        \put(8,12){$d_y$}
        \put(9.5,8){$d_x$}

        \put(69,8.5){$d_y$}
        \put(71,4.5){$d_x$}

        \put(13,30){$d_1$}
        \put(26,24){$d_2$}
        \put(3,12){$d_4$}
        \put(15,5){$d_3$}
        
    \end{overpic}
    \vspace{0.3cm}
    \caption{The slicing of the intermediate map. The intermediate map can be very large, so to reduce the computational overhead of loading and processing it, only the required slice of the map is extracted for processing. If the path planning cannot be performed within the extracted map, the $d_y$ and $d_x$ parameters are increased by a factor to extract a new, large slice of the map for path planning (Section\ref{sec:path_kino})}
    \label{fig:extracted_map}
\end{figure*}

The time and spatial complexity of the Hybrid A*\cite{li2016hybrid} algorithm increases exponentially as the size of the intermediate map increases. Therefore, a slice of the intermediate map (Fig.\ref{fig:extracted_map}) is extracted for each identified path segment to reduce the time and spatial complexity of the Hybrid A* algorithm. The sliced map dimension is calculated as follows, given the start $\mathbf{s}^i_o$ and the target $\mathbf{s}^i_o$: 
\begin{equation*}
    \begin{aligned}
d1 = \mathbf{s}^i_o + \vec{\mathbf{v}} \cdot d_y - \vec{\mathbf{u}}\cdot d_x, 
d2 = \mathbf{s}^i_o - \vec{\mathbf{v}}\cdot d_y - \vec{\mathbf{u}}\cdot d_x, \\
d3 = \mathbf{t}^i_o - \vec{\mathbf{v}}\cdot d_y + \vec{\mathbf{u}}\cdot d_x, 
d4 = \mathbf{t}^i_o + \vec{\mathbf{v}}\cdot d_y + \vec{\mathbf{u}} \cdot d_x, \\
    \end{aligned}   
\end{equation*} where $\vec{\mathbf{u}} = \frac{\mathbf{t}^i_o - \mathbf{x}}{|\mathbf{t}^i_o - \mathbf{s}^i_o|}$, and $\vec{\mathbf{v}} = \left[ -\vec{\mathbf{u}}_y \; \vec{\mathbf{u}}_x \right]$. The terms $d_x$ and $d_y$ define the offset distances along the $x$- and $y$-axes, respectively, used to extract a local planning region around the start and target points. To prevent numerical instability when the start point $\mathbf{s}^i_o$ and target point $\mathbf{t}^i_o$ are close to each other, a minimum offset threshold is enforced. Initially, $d_x$ and $d_y$ are set to 100 pixels. If the path planner fails to find a valid solution within the initial local region, these offsets are progressively increased in size until a valid path is found or the expanded region reaches the boundaries of the global map. This adaptive expansion strategy ensures both numerical robustness and computational efficiency during local path planning. Path planning is performed using the Hybrid A* algorithm for each of the sliced maps (Section\ref{sec:hybrid_a_star}). 

The expansion strategy, increases the slice size by a constant factor in the rare edge case where the initial dynamic slice is insufficient for planning. This simple mechanism ensures adequate context for the planner, though such cases are infrequent in practice.

\subsection{Improving the curvature by Hybrid A*}\label{sec:hybrid_a_star}

\begin{figure*}
    \centering
    \begin{overpic}[width=1\linewidth]{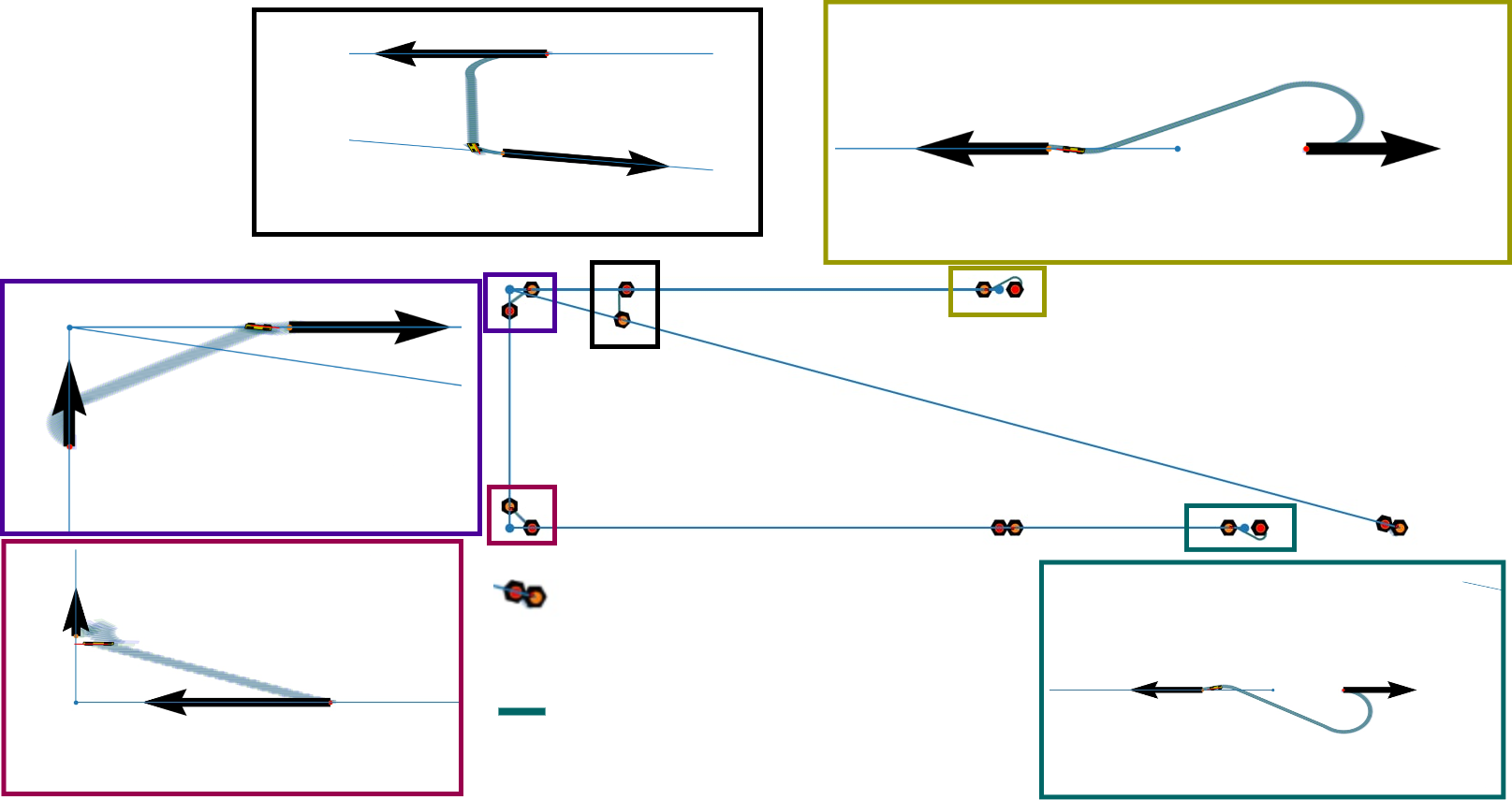}
        \put(37,14){\small Path segments for applying kinodynamic}
        \put(37,12){\small path planning using Hybrid A* }
        \put(37,6){\small Planned path using JPS (or A*) with}
        \put(37,4.5){\small Dijkstra algorithm}
    \end{overpic}
    \vspace{0.5cm}
    \caption{The process of obtaining a kinematically feasible path. The proposed approach uses a hierarchical planning approach, where path planning is performed first, followed by kinematic feasibility checking using the Hybrid A* algorithm. This is because the displacement between the start and target poses can be more than a few kilometres. To do this, the first step is to identify the path segments where kinematic feasibility checking is required (Section.\ref{sec:path_kino}). Afterwards, apply Hybrid A* algorithm}
    \label{fig:kino_path_planning}
\end{figure*}

The process of computing a path that satisfies the vehicle’s kinematic constraints is known as kinematic path planning. In our method, this is achieved using the Hybrid A* algorithm, which is applied to individual segments of the path. For each segment, a dynamically sliced local map is extracted to focus computation on the relevant area. Specifically, once a segment is defined between two intermediate states, $Q_1$ (start state) and $Q_2$ (goal state), Hybrid A* is used to generate a feasible path from $Q_1$ to $Q_2$ within that local map. If a valid path is successfully computed between $Q_1$ and $Q_2$, it is incorporated into the global path. This segment-wise application ensures that the final trajectory is both kinematically feasible and computationally efficient (Fig.\ref{fig:kino_path_planning}). 
\vspace{0.5cm}
 The curvature of the path is improved after applying Hybrid A*-based path planning because the cusps and corners are removed along the identified path segments. 

\subsection{Path length and smoothness improvement}\label{sec:smoothing}

The paths generated using the JPS, A*, or Dijkstra algorithms are not asymptotically optimal\cite{karaman2011sampling}. Therefore, several factors are considered when smoothing the planned path. First, the path must be adjusted to maintain a safe distance from nearby obstacles, whether in narrow or wide passages. Second, while Hybrid A* planning segments may handle curvature well, other segments of the path may experience abrupt changes in curvature that need to be minimized or eliminated during smoothing. Third, JPS or A* might produce only a few intermediate points along the path, so the distance between consecutive points should be minimized. These considerations collectively improve the smoothness of the planned path, as discussed in detail below.

The paths generated by graph-based planners such as JPS, A\*, or Dijkstra are not asymptotically optimal and typically do not account for vehicle kinematic constraints \cite{karaman2011sampling}. To address this limitation, the initial path generated by JPS or A\* is post-processed using a smoothing step that incorporates vehicle kinematics modeled via the bicycle model. Specifically, Hybrid A\* is employed to refine local segments of the path where curvature constraints are critical. This integration ensures that the final trajectory adheres to the vehicle’s non-holonomic motion limits while preserving computational efficiency from the original planner.

Several key factors are considered during this smoothing process. First, the path is adjusted to maintain a safe clearance from nearby obstacles, which is essential in both narrow and wide passages. Second, while Hybrid A\* can handle curvature transitions locally, segments generated by JPS or A\* may include abrupt heading changes that violate kinematic constraints; these are smoothed using curvature-aware optimization consistent with the bicycle model. Third, because graph-based planners like JPS and A\* often produce sparse waypoints, the distance between consecutive points is minimized through interpolation before applying curvature-constrained smoothing.

Together, this approach allows us to leverage the global efficiency of JPS or A\* while ensuring the final path is dynamically and kinematically feasible. Detailed implementation of this hybridization, including curvature computation and trajectory refinement based on the bicycle model, can be added to clarify the integration process.

The obstacle cost is defined using the Voronoi field\cite{lau2010improved}. While similar to the Artificial Potential Fields (APF) method, the Voronoi field offers several advantages over APFs\cite{dolgov2008practical}. Unlike the APF method, which tends to create high-potential zones near narrow passages and may hinder robot navigation through these areas, the Voronoi field adjusts the potential field according to the configuration space's geometry. This adjustment enables robots to navigate through narrow passages more effectively. The obstacle cost is calculated as follows:
\begin{equation}
    \begin{aligned}
        J_o = \sum_{i=0}^{N}V_f(x_i, y_i),
    \end{aligned}
\end{equation} where term N, denoted for the number of points in the planned path. The Voronoi field ($V_f(x_i, y_i)$) is defined as 
\begin{equation}
    \begin{aligned}
       V_f(x_i, y_i) = \Big(\frac{\alpha}{\alpha+d_o(x_i, y_i)}\Big)\cdot\Big(\frac{d_v(x_i,y_i)}{d_o(x_i, y_i) +  d_v(x_i, y_i)}\Big) \\ \cdot \Big(\frac{(d_o(x_i,y_i)-d_{o}^{max})^2}{(d_{o}^{max})^2}\Big),
    \end{aligned}
\end{equation} where term $\alpha >0 \in\mathbb{R}$ control the drop rate for the largest influence range of Voronoi diagram\cite{lau2013efficient}. The maximum allowable distance, denoted $d_{o}^{max} > 0 \in \mathbb{R}$. The function $d_o(\cdot)$ gives the distance to the closest obstacle. Similarly, the function $d_v(\cdot)$ gives the distance to the closest edge of the Voronoi diagram.  

The curvature penalty was defined as follows: 

\begin{equation}\label{eq:cost_k}
    \begin{aligned}
        J_k = \sum_{i=0}^{N-1} (k_i - sign(k_i)\cdot k_{max})^2,
    \end{aligned}
\end{equation} 

where $ \mathbf{k}_i = \Delta {\theta}_i / \Delta \mathbf{x}_i$ determines instantaneous curvature, $sign(\mathbf{k}_i)$  integrates negative and positive curvature assuming that they have the same maximum magnitude evaluated as $k_{max}=1/\rho_{min}$ as shown in Fig.\ref{fig:sign}.  

\begin{figurehere}
    \centering
    \includegraphics[width=\linewidth]{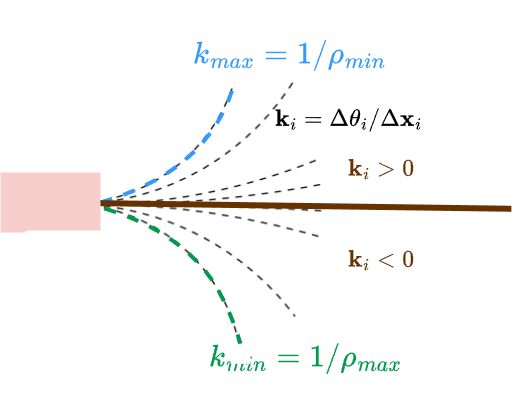}
    \caption{The influence of curvature on turning distance. The curvature is varied in between $1/\rho_{max} \leq k_{i} \leq 1/\rho_{min}$. The turning radius range is defined as $\rho_{min} \leq \rho \leq \rho_{max}$, where $\rho_{max} = |\rho_{min}|$. The turning radius range depends on the kinematic model of the vehicle.}
    \label{fig:sign}
\end{figurehere} 

Here, $\Delta \theta_i = atan2(\Delta \mathbf{y}_{i+1},\Delta \mathbf{x}_{i+1}) - atan2(\Delta \mathbf{y}_{i},\Delta \mathbf{x}_{i})$ corresponds to the change in the steering angle, $\Delta \mathbf{x}_i =  \mathbf{x}_i - \mathbf{x}_{i+1}$, $\Delta \mathbf{y}_i =  \mathbf{y}_i - \mathbf{y}_{i+1}$ denoted the displacements between two consecutive points at $i^{th}$ instance.  The cost for the path length improvement is defined as follows:
 \begin{equation}
    \begin{aligned}
        J_s =\sum_{i=0}^{N-1} 
((\Delta \mathbf{x}_{i+1} - \Delta \mathbf{x}_{i})^2+(\Delta \mathbf{y}_{i+1} - \Delta \mathbf{y}_{i})^2)
    \end{aligned}
\end{equation} 
The term $J_s$ is used to reduce the distance between consecutive points along the planned path. Therefore, the overall objective is defined as follows: 
\begin{equation}
    \begin{aligned}
        J = \lambda_o J_o + \lambda_k J_k + \lambda_s J_s,
    \end{aligned} 
\end{equation} where weights parameters for each of the cost terms are denoted by $\lambda_*, \; * \in o, k, s$ whose values were set as 0.3, 0.4, and 0.3, which were estimated empirically.
\begin{figure*}[t]
    \centering    
     \begin{overpic}[width=1\linewidth]{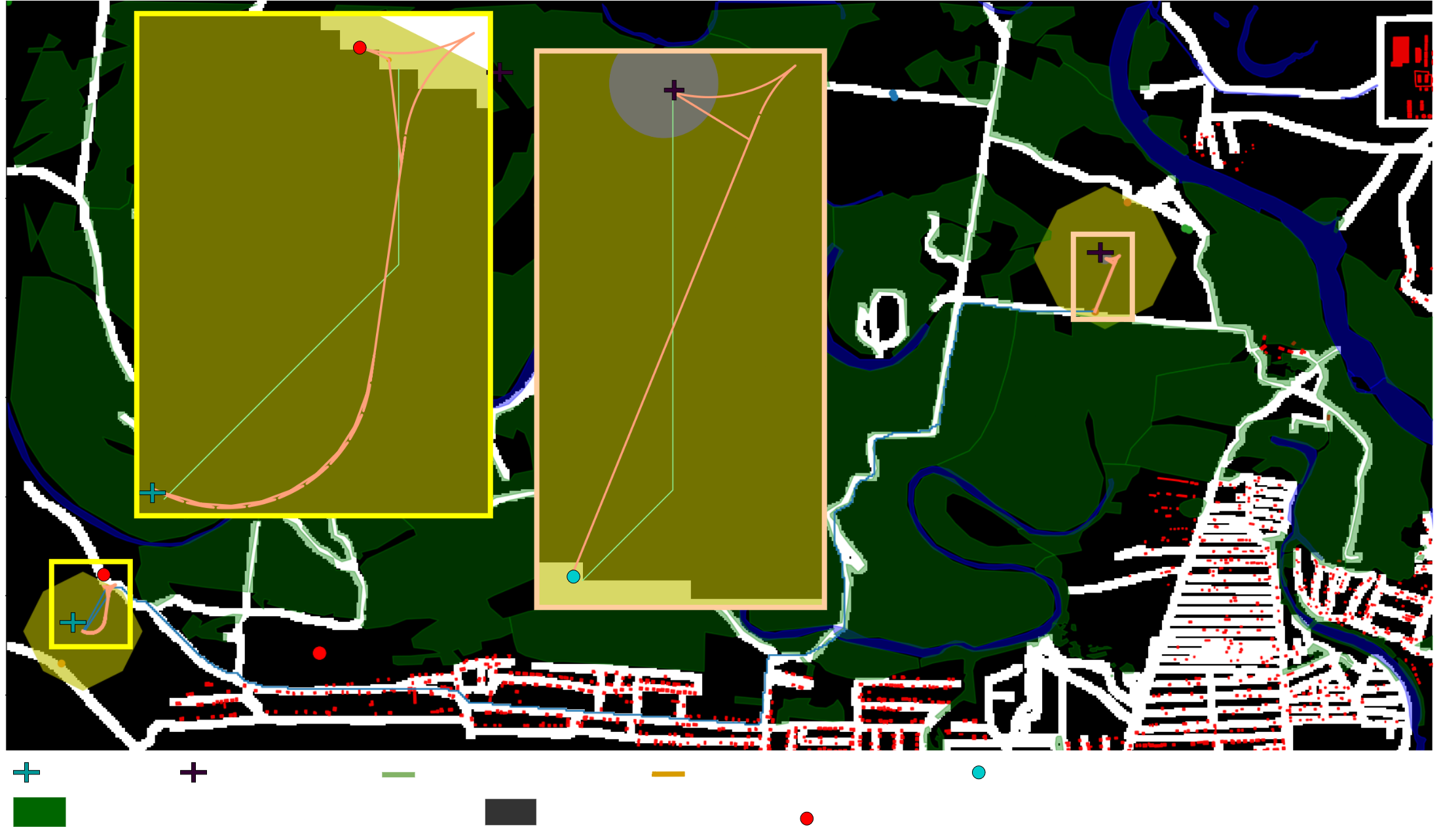}
        \put(3,4.3){\small Start pose}
        \put(15,4.3){\small Target pose}
        \put(29,4.3){\small Initial planned path}
        \put(48,4.3){\small Smoothed path segments}
        \put(68.5,4.3){\small Closest target pose on the off-road trails}
        \put(5,1.5){\small Occupied area}
        \put(38,1.5){\small Traversable area}
        \put(57,1){\small Desired closest start pose on the off-road trails}
    \end{overpic}

    \caption{The path smoothing technique introduced in Section~\ref{sec:smoothing} is designed to refine the path generated by the Hybrid A* algorithm. Hybrid A* typically produces a feasible but non-smooth path, especially around the connecting joints where transitions occur between off-road trails and the start or target pose. To address this, the proposed smoothing technique is applied specifically to these segments—between the connecting joints and the desired start or goal pose, ensuring a smoother and more navigable trajectory for the vehicle.}
    \label{fig:smoothing}
\end{figure*} 
An example scenario of the proposed path length and smoothness improvement is shown in Fig.~\ref{fig:smoothing}. 
\subsection{Passable and restricted areas}
Both passable and restricted areas can be defined through a Graphical User Interface (GUI). These areas can be convex or concave. Hence, the winding number algorithm~\cite{alciatore1995winding} is utilized for finding the area that belongs to a specified shape within the intermediate map and updating it accordingly. Specifically, user-restricted areas are assigned as obstacles, and user-passable areas are assigned as free spaces. 

\begin{figurehere}
    \centering
    \begin{overpic}[width=1\linewidth]{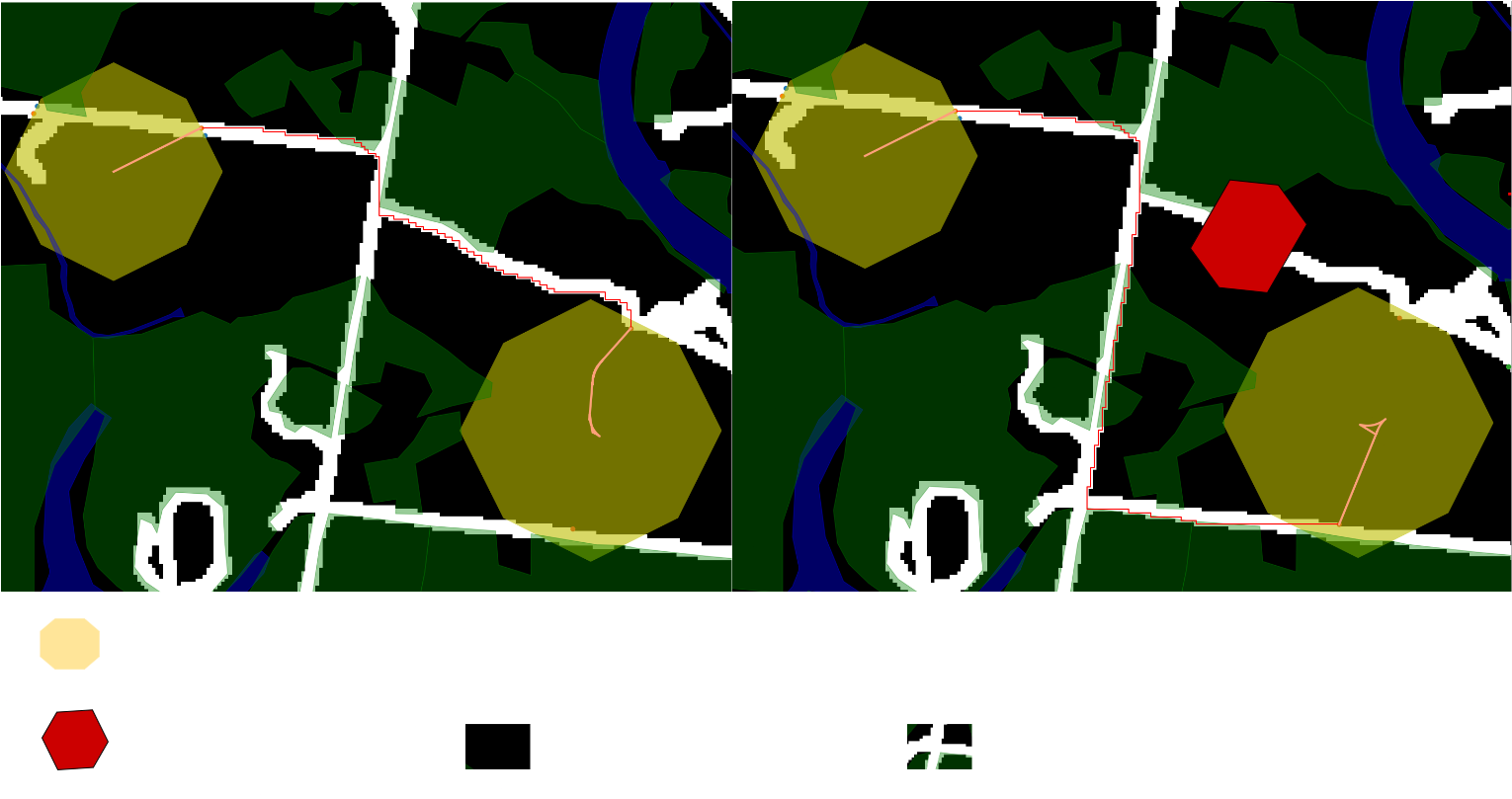}
        \put(0,54){\small Before adding restricted areas }
        \put(50,54){\small After adding restricted areas }

        \put(7, 2){\small Restricted}
        \put(7, 9){\small Polygon to identify poses nearest to the goal location.}
        \put(35, 2){\small Traversable}
        \put(65, 2){\small Road network}
    \end{overpic}
    \caption{An example of path planning, both before and after incorporating user-defined restrictions alongside the intermediate map.}
    \label{fig:passible_area}
\end{figurehere}

In Fig.~\ref{fig:passible_area}, it is shown that behaviours of the path planning when introducing these area types. 

\section{Software Framework Validation and Results}
\begin{figurehere}
    \centering
    \includegraphics[width=0.9\linewidth]{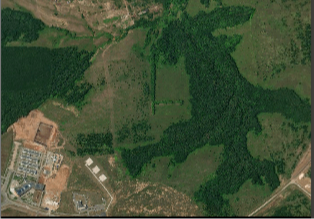}
    \caption{The second map, referred to as Map 2, was utilized to validate the proposed system. Map 2 was generated through a 3D reconstruction of the environment.}
    \label{fig:inn_map}
\end{figurehere} 

To illustrate and evaluate the performance of the proposed software framework, several different types of experiments were carried out. 

\begin{table*}
\tbl{A comparison of the average computation time and average memory utilization for different off-road global planning methods. Kinematic feasibility is of utmost importance in off-road global planning because it ensures that the obtained path is traversable by the vehicle. A path planner that is characterized as online can handle real-time processing tasks.\label{tab:comparison}}
{\begin{threeparttable}
\small 
\resizebox{\textwidth}{!}{%
\begin{tabular}{|l|c|c|l|l|c|c|l|l|c|}
\hline
\multirow{2}{*}{Method} & \multicolumn{8}{c|}{Map 1} & \multirow{2}{*}{\begin{tabular}[c]{@{}c@{}}Online \\ or \\ Offline\end{tabular}} \\
\cline{2-9}
 & \begin{tabular}[c]{@{}c@{}}Time\\ (s)\end{tabular} & \begin{tabular}[c]{@{}c@{}}Memory\\ (GB)\end{tabular} & \begin{tabular}[c]{@{}c@{}}CSD\\ (m$^{-1}$)\end{tabular} & \begin{tabular}[c]{@{}c@{}}MOD\\ (m)\end{tabular} & \begin{tabular}[c]{@{}c@{}}Time\\ (s)\end{tabular} & \begin{tabular}[c]{@{}c@{}}Memory\\ (GB)\end{tabular} & \begin{tabular}[c]{@{}c@{}}CSD\\ (m$^{-1}$)\end{tabular} & \begin{tabular}[c]{@{}c@{}}MOD\\ (m)\end{tabular} &  \\
\hline
 & \multicolumn{4}{c|}{Without road network} & \multicolumn{4}{c|}{With road network} & \\
\hline
\begin{tabular}[c]{@{}l@{}}A*~\cite{kulathunga2020real} \\ \textit{deterministic,} \\ \textit{confined} \\ \textit{exploration} \end{tabular} & 1.35 & 1.25 & 0.25 & 1.0 & 0.74 & 0.95 & 0.16 & 1.0 & Online \\
\rowcolor[rgb]{0.945,0.596,0.788}
\begin{tabular}[c]{@{}l@{}}JPS~\cite{liu2017planning}, \\ \textit{deterministic}, \\ \textit{confined} \\ \textit{exploration} \end{tabular} & 1.14 & 1.12 & 0.23 & 1.0 & 0.56 & 0.85 & 0.14 & 1.0 & Online \\
\begin{tabular}[c]{@{}l@{}}RRT*~\cite{kulathunga2021path} \\ probabilistic \\ unbounded search \end{tabular} & Inf & Inf & Inf & Inf & Inf & Inf & Inf & Inf & Offline \\
\begin{tabular}[c]{@{}l@{}}Informed RRT* \\ probabilistic \\ restricted search \end{tabular} & Inf & Inf & Inf & Inf & Inf & Inf & Inf & Inf & Offline \\
\rowcolor[rgb]{0.749,0.741,0.945}
\begin{tabular}[c]{@{}l@{}}The proposed \\ \textit{\textbf{kinematically}} \\ \textit{\textbf{feasible,}} \\ \textit{\textbf{deterministic,}} \\ \textit{\textbf{memory-efficient}}\end{tabular} & 1.69 & 1.28 & 0.09 & 1.6 & 0.89 & 1.14 & 0.08 & 1.8 & Online \\
\hline
\multicolumn{10}{|c|}{\textbf{Map 2}} \\
\hline
 & \multicolumn{4}{c|}{Without road network} & \multicolumn{4}{c|}{With road network} & \\
\hline
A*~\cite{kulathunga2020real} & 1.10 & 0.81 & 0.35 & 1.0 & 0.45 & 0.23 & 0.26 & 1.0 & Online \\
\rowcolor[rgb]{0.945,0.596,0.788}
JPS~\cite{liu2017planning} & 0.90 & 0.68 & 0.31 & 1.0 & 0.32 & 0.14 & 0.20 & 1.0 & Online \\
RRT*~\cite{kulathunga2021path} & Inf & Inf & Inf & Inf & Inf & Inf & Inf & Inf & Offline \\
Informed RRT* & Inf & Inf & Inf & Inf & Inf & Inf & Inf & Inf & Offline \\
\rowcolor[rgb]{0.749,0.741,0.945}
Proposed & 1.39 & 0.90 & 0.09 & 1.4 & 1.14 & 0.56 & 0.04 & 1.3 & Online \\
\hline
\end{tabular}%
}
\begin{tablenotes}
  \item[1] \textbf{Inf} denotes planning algorithm fails to find a feasible path within ten seconds, \textbf{GB} Gigabyte, \textbf{CSD} curvature standard deviation, \textbf{MOD} minimum obstacle distance
\end{tablenotes}
\end{threeparttable}}
\end{table*}

\subsection{System setup}
We conducted all the experiments on a computer with these specifications: Intel(R) Core(TM) i5-8265U CPU and 16GB of RAM. The Python programming language was used for the preprocessing and generation of the initial intermediate map. The proposed framework was written in C++, adhering to the C++17 specifications. We used Ubuntu as the operating system to run the framework. Consequently, the framework is not tied to a specific operating system.
\subsection{Input maps specifications}
We used two different maps for all the experiments described in the following sections. Map 1 had a resolution of 5000 x 16000, and Map 2 (Fig.\ref{fig:inn_map}) had a resolution of 6000 x 4000. Both maps were converted into the pixel coordinate system before the experiments were conducted.
Both maps cover several square kilometres. Both Map 1 and Map 2 consist of several geographical features, including trees, off-road trails, rivers, bodies of water, and user-restricted areas. 

\begin{figure*}[ht!]
    \centering
     \begin{overpic}[width=1\linewidth]{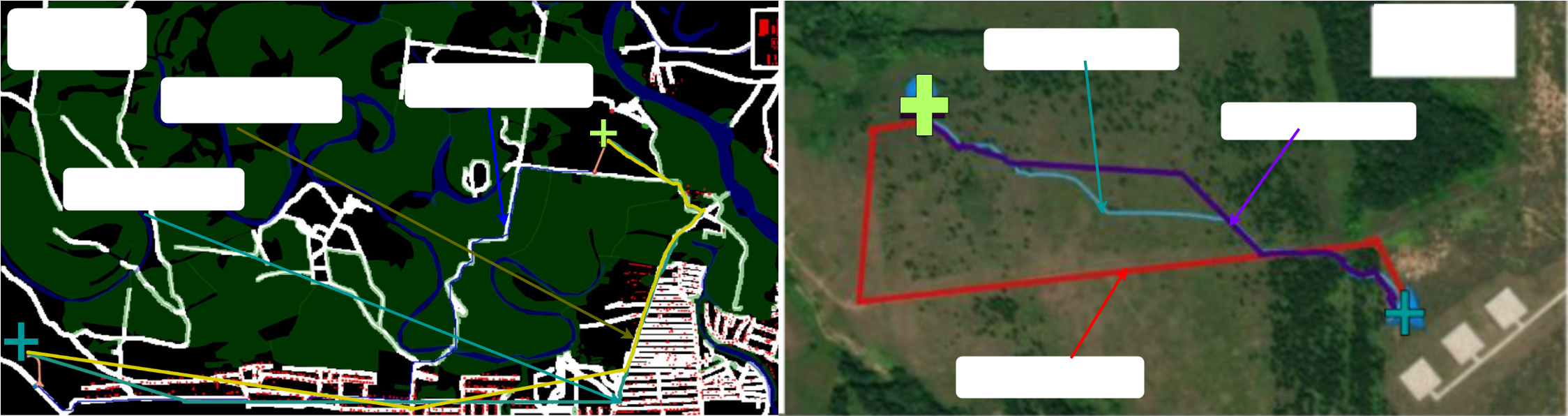}
        \put(2,23.5){Map 1}
        \put(89,23.5){Map 2}
        \put(4,14){\color[HTML]{009999} A* algorithm}
        \put(9.5,19.5){ \color[HTML]{666600} JPS algorithm}
        \put(25,20.5){ \color[HTML]{0000FF} The proposed}

        \put(61,2){ \color[HTML]{FF0000} A* algorithm}
        \put(77.5,18.3){ \color[HTML]{7F00FF} JPS algorithm}
        \put(63,23){ \color[HTML]{009999} The proposed}
     \end{overpic}
    \caption{An illustration of the proposed approach for off-road global path planning is shown for both Map 1 and Map 2.  For Map 2, we constructed a 3D map of the environment using point clouds and projected the planning onto a latitude-longitude coordinate system.}
    \label{fig:global_planning}
\end{figure*}

\subsection{Average execution time and memory utilization}\label{sec:execution}

To assess the performance of our off-road navigation framework, we measured two key metrics: average execution time and memory utilization for global path planning. Average execution time is the average execution time it takes for global planning. Memory utilization is the amount of memory that is utilised for global planning. We used Procpath profiler \cite{procpath} to track these metrics. The intermediate map was constructed in the pixel coordinate system, so we can easily deploy any planning technique for validation purposes. For a comprehensive comparison, two graph search-based methods: A*\cite{kulathunga2020real} and JPS\cite{liu2017planning}, and two sampling-based methods: RRT*\cite{kulathunga2021path} and Informed RRT* were used. These metrics are critical for evaluating the efficiency and practicality of each algorithm, particularly in terms of kinematic feasibility and real-time processing capabilities. Initially, we selected 100 different start and target poses in each map, ensuring that the minimum displacement between any two start and target points is at least 1 kilometer (Fig.\ref{fig:global_planning}).

The A* algorithm demonstrates relatively low average computation times and memory usage across both maps, making it suitable for real-time applications as shown in Table \ref{tab:comparison}. Specifically, on Map 1, A* requires 1.3567 seconds and 1.25 GB of memory, while on Map 2, it takes 0.7467 seconds and 0.95 GB of memory. Its deterministic and confined exploration approach contributes to its efficiency, though it may not handle complex kinematic constraints as effectively as other methods. In comparison, the JPS (Jump Point Search) algorithm shows slightly better performance than A* in terms of both computation time and memory usage. For Map 1, JPS requires 1.1459 seconds and 1.12 GB of memory, and for Map 2, it takes 0.5641 seconds and 0.85 GB of memory. JPS, like A*, is deterministic and employs confined exploration, but its enhancements allow for more efficient pathfinding and memory usage.

On the other hand, RRT* and Informed RRT* algorithms exhibit significant limitations. Both fail to produce feasible paths within the ten-second limit, as indicated by their "Inf" values for computation time and memory usage. This failure suggests that these probabilistic algorithms, whether unbounded or restricted in their search strategies, are not suitable for real-time applications in their current forms.

The proposed method presents a balanced trade-off between efficiency and kinematic feasibility. It shows a moderate increase in computation time and memory usage compared to A* and JPS, but it provides crucial benefits in terms of kinematic feasibility and real-time capabilities. For Map 1, this method requires 1.6990 seconds and 1.28 GB of memory, and for Map 2, 0.8901 seconds and 1.14 GB of memory. While less efficient than JPS and A*, it compensates by offering a solution that meets the kinematic constraints essential for complex off-road scenarios.

\begin{figurehere}
    \centering
    \includegraphics[width=\linewidth]{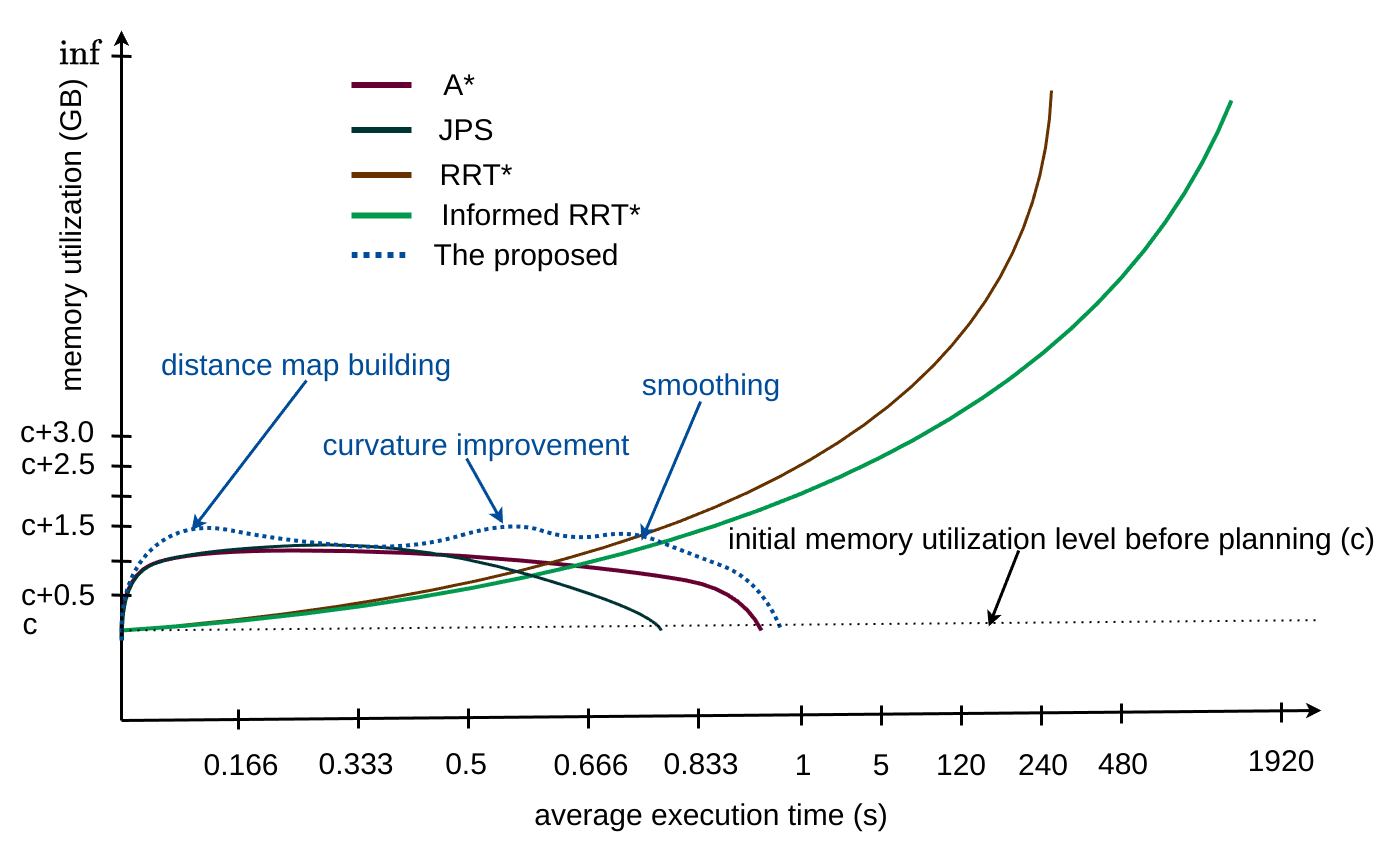}
    \caption{The average computation time and average memory utilization for off-road global planning were compared. The results were based on planning on Map 1, as described in Section~\ref{sec:execution}. The proposed approach took a slightly higher average execution time than JPS and A*, but it also considered smoothing and kinematic feasibility checking, which the other two approaches did not.}
    \label{fig:time_execution_all}
\end{figurehere}

In summary, the analysis reveals that A* and JPS are both efficient online algorithms suitable for real-time applications, with JPS outperforming A* in terms of computation time and memory usage. However, 
The proposed approach prioritizes kinematic feasibility while maintaining reasonable performance: its average execution time and memory utilization are slightly higher than the graph search methods (Fig.\ref{fig:time_execution_all}), it guarantees a collision-free, vehicle-maneuverable path (unlike A* and JPS). 
This trade-off is achieved through these additional operations: distance map building that provides efficient path cost estimation, hybrid A* for curvature improvement that ensures smoother, kinematically feasible paths, and path smoothing that further refines the path for feasible traversal.
Through these additional operations, our framework guarantees kinematic feasibility while maintaining acceptable real-time performance. The choice of algorithm should be based on the specific needs of the application, including real-time processing requirements and the ability to handle kinematic constraints. 
Additionally, in Figure \ref{fig:time_execution_all}, the horizontal axis employs an approximate logarithmic scale to accommodate the wide range of execution times (0.166 to 1920 seconds). This prevents clustering of short execution times and enables clear visual comparison across all data points. 

\section{Discussion and Conclusion}
The proposed framework's performance is sensitive to extreme terrain conditions, such as dense vegetation, steep slopes, and rocky environments, which can affect LiDAR data quality and map accuracy. LiDAR penetration is limited in areas with dense foliage, potentially leading to incomplete point clouds and misclassifications of traversable regions. While ground filtering and vegetation classification help mitigate these issues, sensor limitations and environmental factors can still introduce inaccuracies. Additionally, GPS inaccuracies in areas with poor satellite visibility may affect georeferencing. To address these challenges, sensor fusion with IMU data or multi-spectral imagery can be employed to enhance robustness, particularly in difficult terrains.

For ground-based other vehicles, the path planning approach can accommodate other non-holonomic motion models for wheeled robots, ensuring feasible paths considering vehicle dynamics. For tracked or legged platforms, additional terrain traversability metrics, such as slope gradients and surface roughness, can be integrated. Further work could involve incorporating vehicle-specific cost functions into the path planner to optimize paths for different mobility platforms, expanding the system's applicability across both aerial and ground vehicles in off-road environments.

The proposed off-road global planner addresses essential factors beyond the traditional goals of minimizing path length and travel time, including real-time performance, kinematic feasibility, and efficient memory utilization. This framework successfully balances these objectives by focusing on both memory efficiency and average computation time while maintaining the core goal of optimal pathfinding. The integration of preprocessing functionalities and intermediate map generation based on provided geographical features further enhances the practicality of the proposed method. A key area for future work will be the development of local path planning strategies that address traversability within unstructured environments. While the global planner efficiently manages large-scale path planning, incorporating local path planning into the framework will enable more detailed and precise navigation over shorter horizons. This step is crucial for improving the robot's ability to handle dynamic and complex terrains where real-time adaptability is required. 

Moreover, while the current focus has been on optimizing memory usage and computation time, future research could explore advanced techniques to further reduce these parameters without compromising the quality of path planning. Future work could also benchmark the computational overhead introduced by the Voronoi field adjustment and curvature penalty in real-time scenarios. Evaluating their runtime impact across different map resolutions, obstacle densities, and hardware optimizations will provide insights into their efficiency. These analyses will help refine the trade-off between path quality and computational cost in off-road navigation. Additionally, addressing the challenge of traversability in unstructured environments—such as those with sudden changes in terrain or environmental conditions—will be crucial. Implementing a robust approach to local path planning will require balancing real-time performance with the demands of high memory utilization and computation time, as these factors often interact in complex ways. Furthermore, we have already demonstrated the practical integration of the proposed global planner with a local planner in our recent work \cite{jerome2025genetic}, where the global planner provides long-range guidance and the local planner ensures kinodynamic feasibility in complex, uneven terrains.

While the proposed method shows strong potential, it is important to acknowledge its current limitations. The planner operates under the assumption of a static environment and does not account for dynamic obstacles such as moving vehicles, animals, or changing terrain conditions. Additionally, the method faces challenges in extremely complex or unstructured terrains, such as dense vegetation, cliffs, or uneven surfaces, where accurate feature extraction and traversability assessment become more difficult. These limitations were evident in some of our experimental results and are important considerations for practical deployment. Ultimately, integrating dynamic obstacle handling, more robust terrain modeling, and adaptive traversability analysis will lead to a more comprehensive solution for autonomous navigation in off-road settings. Such improvements will potentially extend the framework’s applicability to more diverse and challenging environments. Future work will focus on refining these components and validating their performance in real-world scenarios, ensuring the planner remains both practical and effective across a broader range of conditions.

\bibliographystyle{ws-us}
\bibliography{references}

\noindent\includegraphics[width=1in]{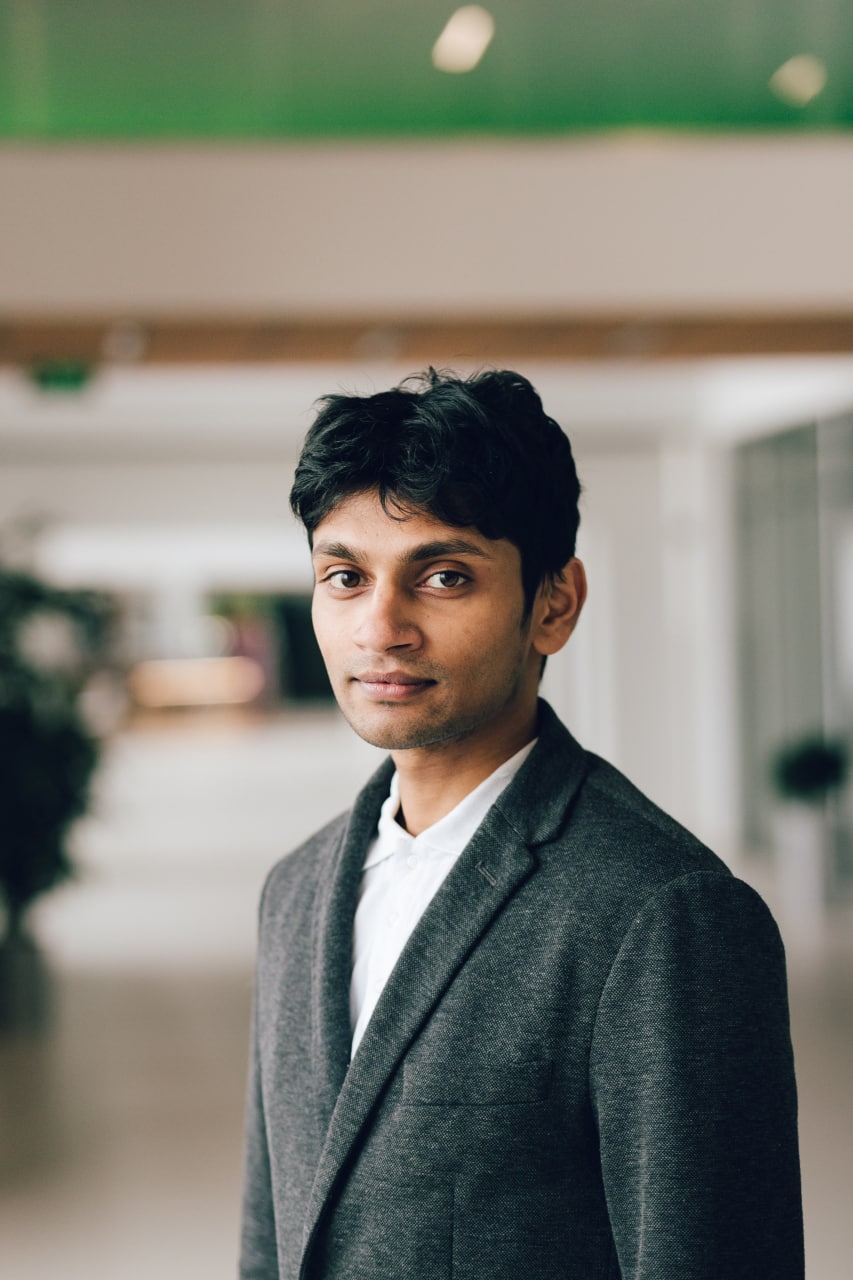}  
{\bf Geesara Kulathunga} received his B.Eng(Hons). degree in computer engineering from Peradeniya University, Sri Lanka, in 2015, and his Ph.D. degree in computer science and engineering from Innopolis University, Kazan, Russia. He is currently a postdoctoral researcher at the Faculty of Computer Science of the University of Lincoln, UK.  
\\
\noindent\includegraphics[width=1in]{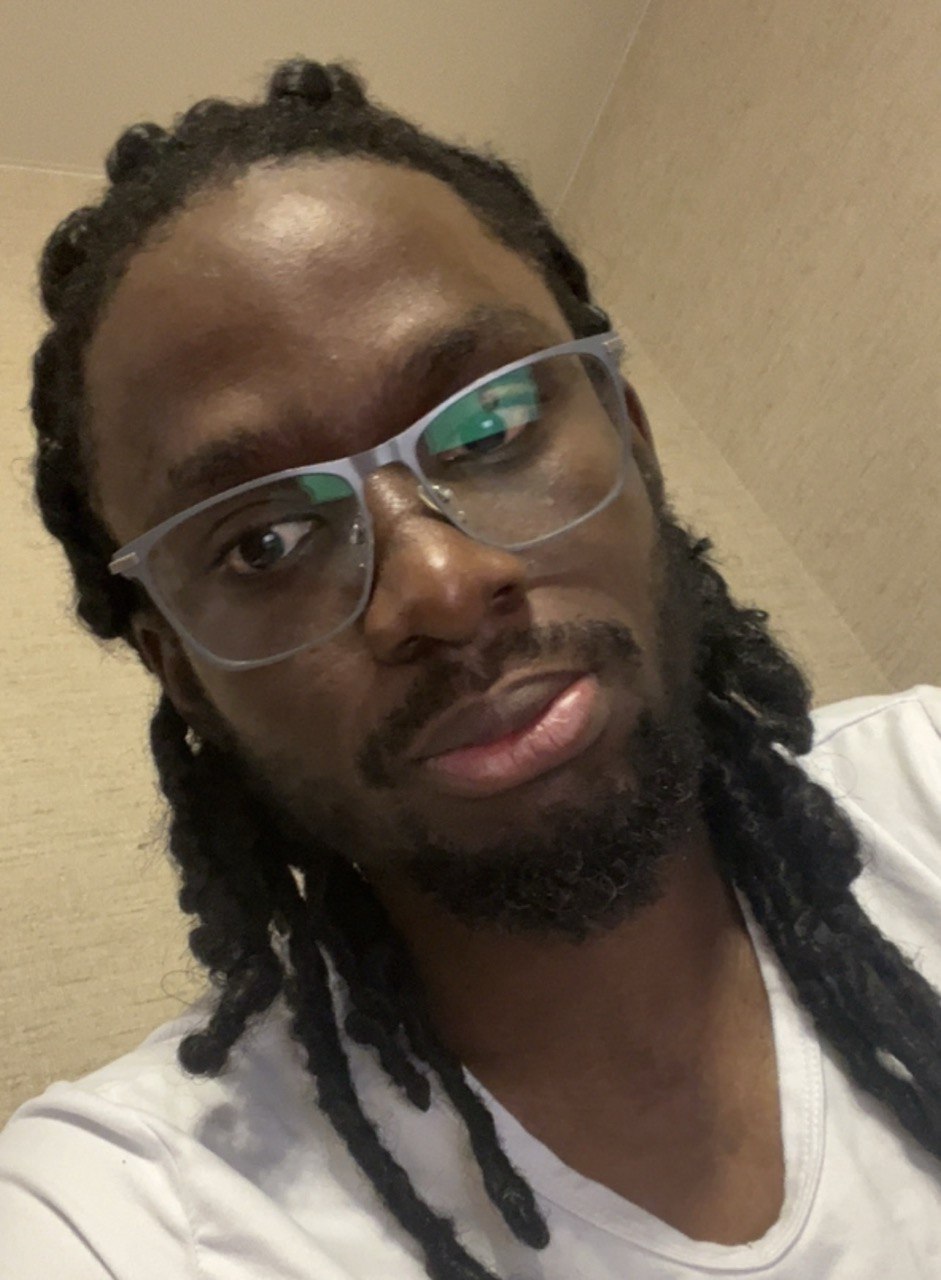}  
{\bf Otobong Jerome} received his B.Eng(Hons). degree in computer engineering from the University of Uyo, Nigeria, and his M.Sc. degree in Robotics/Computer Vision from Innopolis University, Innopolis, Russia. He is currently a Ph.D. candidate in Mathematical Modelling and Numerical Methods at Innopolis University, Russia. Otobong works at the Center for Unmanned Aerial Vehicles, where he is responsible for research projects in the area of autonomous systems.  
\\
\noindent\includegraphics[width=1in]{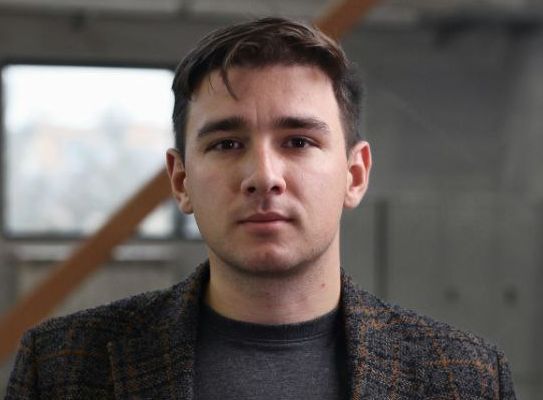}  
{\bf Devitt Dmitry} received his B.Eng(Hons). degree in mechatronic and robotic engineering from Southern Federal University, Russia, in 2017, and his Ph.D. degree in computer science and engineering from Innopolis University, Kazan, Russia. He is currently the head of the unmanned aerial systems center at Innopolis University, Russia.  
\\
\noindent\includegraphics[width=1in]{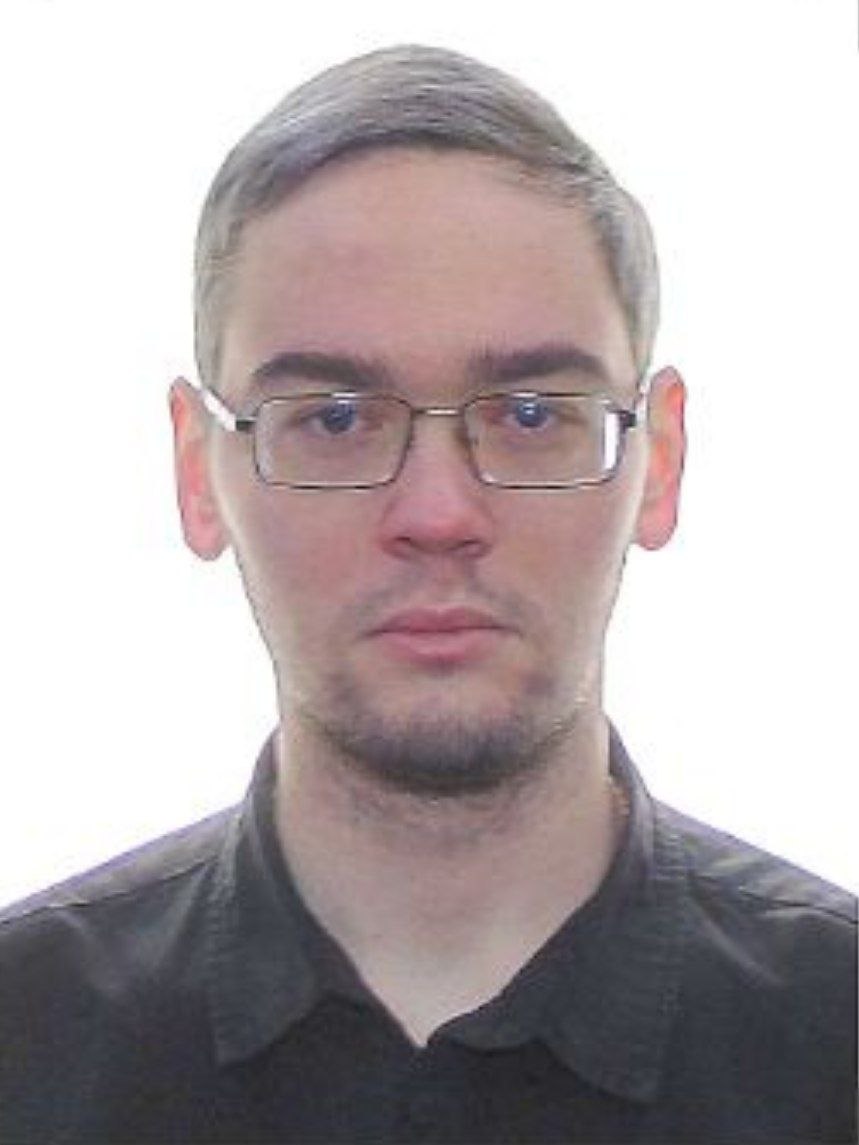}  
{\bf Evgeniy Muravev} received his B.Eng(Hons). degree in computer science and engineering from Innopolis University, Russia, in 2021. He is currently employed at the unmanned vehicle systems center at Innopolis University, Russia.  

Evgeniy Muravev is the author of over 3 technical publications. His research interests include computer science applications in unmanned vehicle systems and autonomous technologies.
\end{multicols}
\end{document}